%% file: main.tex
\documentclass[sigconf]{acmart}
\usepackage{algorithmic}
\usepackage{graphicx}
\usepackage{textcomp}
\usepackage{xcolor}
\usepackage{fancyhdr}
\usepackage[skins,theorems]{tcolorbox}
\usepackage[linesnumbered, ruled, boxed]{algorithm2e}
\usepackage{enumitem}
\usepackage{booktabs}
\usepackage{setspace}
\AtBeginDocument{%
  \providecommand\BibTeX{{%
    \normalfont B\kern-0.5em{\scshape i\kern-0.25em b}\kern-0.8em\TeX}}}

\setcopyright{acmcopyright}
\copyrightyear{2018}
\acmYear{2018}

\acmConference[PACT '22]{PACT '22:  International Conference on
Parallel Architectures and Compilation Techniques (PACT)}{October 10--12, 2022}{Chicago,IL}
\acmBooktitle{PACT '22: International Conference on
Parallel Architectures and Compilation Techniques (PACT), October 10--12, 2018, Chicago, IL}
\acmPrice{15.00}
\acmISBN{978-1-4503-XXXX-X/18/06}

\input{macros.tex}

\begin{document}
\input{tex/abstract.tex}

\title{\rev{GNNear}: Accelerating Full-Batch Training \rev{of Graph Neural Networks} with Near-Memory Processing }




\author{Zhe Zhou}
\authornotemark[1]
\affiliation{%
  \institution{School of Integrated Circuits}
    \institution{School of Computer Science}
        \institution{Peking University}
  \city{Beijing}
  \country{China}}
\email{zhou.zhe@pku.edu.cn}

\author{Cong Li}
\authornotemark[1]
\affiliation{%
    \institution{School of Integrated Circuits}
        \institution{Peking University}
  \city{Beijing}
  \country{China}}
\email{leesou@pku.edu.cn}

\author{Xuechao Wei}
\affiliation{%
    \institution{School of Computer Science}
        \institution{Peking University}
  \institution{Alibaba Group}
  \city{Beijing}
  \country{China}}
\email{xuechao.wei@pku.edu.cn}

\author{Xiaoyang Wang}
\affiliation{%
    \institution{School of Integrated Circuits}
    \institution{School of Computer Science}
        \institution{Peking University}
  \city{Beijing}
  \country{China}}
\email{yaoer@pku.edu.cn}

\author{Guangyu Sun}
\authornotemark[2]
\affiliation{%
    \institution{School of Integrated Circuits}
    \institution{Peking University}
        \institution{Beijing Advanced Innovation Center for Integrated Circuits}
  \city{Beijing}
  \country{China}}
\email{gsun@pku.edu.cn}

\renewcommand{\shortauthors}{Zhe and Cong, et al.}



\maketitle

\input{tex/introduction_2.tex}
\input{tex/background.tex}

\input{tex/motivation.tex}

\input{tex/architecture.tex}

\input{tex/system.tex}

\input{tex/evaluation.tex}

\input{tex/discussion.tex}
\input{tex/related_work.tex}
\input{tex/conclusion.tex}


\bibliographystyle{ACM-Reference-Format}
\bibliography{refs}

\end{document}

%% file: macros.tex
\newcommand\rev[1]{\textcolor{black}{#1}}
\newcommand{\myfont}{\fontsize{9pt}{\baselineskip}\selectfont}

\usepackage{titlesec}
\makeatletter
\g@addto@macro{\normalsize}{%
  \setlength{\abovedisplayskip}{3pt plus 0.5pt minus 1pt}
  \setlength{\belowdisplayskip}{3pt plus 0.5pt minus 1pt}
  \setlength{\abovedisplayshortskip}{0pt}
  \setlength{\belowdisplayshortskip}{0pt}
  \setlength{\intextsep}{4pt plus 1pt minus 1pt}
  \setlength{\textfloatsep}{4pt plus 1pt minus 1pt}
  \setlength{\skip\footins}{5pt plus 1pt minus 1pt}
  }
\makeatother

\titlespacing\section{0pt}{2pt plus 1pt minus 1pt}{3pt plus 1pt minus 2pt}
\titlespacing\subsection{0pt}{2pt plus 1pt minus 1pt}{3pt plus 1pt minus 2pt}
\titlespacing\subsubsection{0pt}{2pt plus 1pt minus 1pt}{3pt plus 1pt minus 2pt}

\newcommand{\ogbnUpdateComputation}{100 TFLOPs}

\newcommand{\geomeanSpeedupOverCPU}{$30.8\times$}
\newcommand{\geomeanSpeedupOverGPU}{$2.5\times$}
\newcommand{\geomeanEnergySavingOverCPU}{$79.6\times$}
\newcommand{\geomeanEnergySavingOverGPU}{$7.3\times$}



%% file: tex/abstract.tex
\begin{abstract}
\noindent Recently, \rev{Graph Neural Networks (GNNs)} have become state-of-the-art algorithms for analyzing non-euclidean graph data. However, to realize efficient \rev{GNN} training is challenging,  especially on large graphs. The reasons are many-folded: 1) \rev{GNN} training incurs a substantial memory footprint. Full-batch training on large graphs even requires hundreds to thousands of gigabytes of memory. 2) \rev{GNN} training involves both memory-intensive and computation-intensive   operations, challenging current CPU/GPU platforms. 3) The irregularity of graphs  can result in severe resource under-utilization and load-imbalance problems.  

This paper presents a \texttt{\rev{GNNear}} accelerator
to tackle these challenges.  
\texttt{\rev{GNNear}} adopts a DIMM-based memory system  to provide sufficient memory capacity. To match the heterogeneous nature of \rev{GNN} training, we 
offload the memory-intensive \texttt{Reduce} operations to in-DIMM Near-Memory-Engines (NMEs), making full use of the  high aggregated local bandwidth. We adopt a Centralized-Acceleration-Engine (CAE) to process the computation-intensive \texttt{Update} operations. We further propose several optimization strategies to deal with the irregularity of input graphs and improve \texttt{\rev{GNNear}}'s performance. Comprehensive evaluations on 16 \rev{GNN} training tasks demonstrate that \texttt{\rev{GNNear}} achieves \geomeanSpeedupOverCPU~/~\geomeanSpeedupOverGPU~geomean speedup and \geomeanEnergySavingOverCPU~/ \geomeanEnergySavingOverGPU~(geomean) higher energy efficiency  compared to Xeon E5-2698-v4 CPU and NVIDIA V100 GPU.

\end{abstract}

%% file: tex/introduction_2.tex
\section{Introduction}
\renewcommand{\thefootnote}{\fnsymbol{footnote}}
\footnotetext[1]{These authors contributed equally to this work.}
\footnotetext[2]{Corresponding author.}
\noindent The past few years have witnessed the explosion of deep-learning techniques. To process traditional euclidean data such as 2D images, Convolutional Neural Networks (CNNs) have been invented and become the de-facto tools on diverse tasks such as image classification~\cite{vgg, alexnet,resnet,mobilenets,densenet}, object detection~\cite{yolo,girshick2015fast,ssd,fasterrcnn}, and image segmentation~\cite{segmentation,unet,bisenet, unetpp}. However, though CNNs are powerful in image processing, they cannot process the equally pervasive non-euclidean graph data. Thus,  
there has been an increasing interest in  \rev{Graph Neural Networks~(GNN}s).
We notice that GNNs  have contributed to many breakthroughs on various graph-based tasks, such as  node classification~\cite{Garcia-DuranN17,GibertVB12,semigcn}, point-cloud analysis~\cite{wang2019graph,xu2020grid,qian2021pu,zhai2020multi}, recommendation systems~\cite{ying2018graph,zhao2019intentgc,wu2019session,wang2019kgat,fan2019metapath}, smart traffic~\cite{zhao2019t,ge2019temporal,cui2019traffic}, IC design~\cite{wang2020gcn,mirhoseini2020chip,zhang2019circuit}, physical system simulation~\cite{henrion2017neural}, and drug discovery~\cite{lo2018machine,stokes2020deep}.

Apart from algorithmic  innovations,  there are also some \rev{GNN} accelerators designed to accelerate the inference of various \rev{GNN} algorithms~\cite{hygcn, geng2019awb, blockgnn, liang2020engn, rubik, legognn,gcnx,song2021cambricon,wang2020gnn, geng2021gcn}. However, due to many challenges, neither existing accelerators nor GPU/CPU platforms can easily support \rev{GNN} training at scale. Firstly, \rev{GNN} training incurs a large memory footprint. To enable back-propagation, we have to buffer a huge amount of intermediate data. Even the middle-scale Reddit dataset~\cite{graphSAGE} containing 114 million edges will take up 58 GB of  memory~\cite{fuseGNN} with PyG framework~\cite{pyg}, exceeding  most GPUs' capacity. Meanwhile, as illustrated in Figure~\ref{fig:trend}, the scale of  graphs applied in \rev{GNN} researches has been growing exponentially in recent years~\cite{semigcn,graphSAGE,clustergcn,hu2020ogb}.  Although  mini-batch training strategies leveraging  neighbor-sampling~\cite{fastgcn,clustergcn,graphSAGE,zeng2019graphsaint} successfully  reduce the memory requirement by limiting the training batch size, they are proved to have lower  accuracy compared to full-batch training due to approximation errors~\cite{graphSAGE,distgnn,roc,tripathy2020reducing,dgcl}.
\input{figtex/fig_trend}

Secondly, \rev{GNN} training has both memory-intensive (e.g., 0.5 Ops/Byte) features/gradients reduction and  computation-intensive (e.g., 128 Ops/Byte) features/gradients update operations. Neither PIM/NMP-based graph processing accelerators\cite{zhuo2019graphq,graphp,ahn2015scalable, dai2018graphh,graph_pim} (optimized for memory access)  nor traditional DNN accelerators\cite{chen2014diannao,jouppi2017datacenter,eyeriss, dadiannao, tangram, zhang2015optimizing} (optimized for computation) can handle such a heterogeneous nature in \rev{GNN} training.  Thirdly, the  irregularity of real-world graphs 
will result  in severe resource  under-utilization and load-imbalance problems~\cite{hygcn, geng2019awb, geng2021gcn}, increasing the difficulty of optimizing a \rev{GNN} training system's throughput.

To tackle these challenges, we propose \texttt{\rev{GNNear}}, which harnesses both near-memory processing (NMP) and centralized processing  to achieve high-throughput, energy-efficient, and scalable \rev{GNN} training on large graphs. Specifically,  we first analyze the \rev{GNN} training procedure and categorize the involved operations as a sequence of memory-intensive \texttt{Reduce} and computation-intensive \texttt{Update} operations. We then 
adopt DIMM-based Near-Memory Engines (NMEs) and a powerful Centralized-Acceleration-Engine (CAE)  to process them respectively.  Such a hybrid architecture perfectly matches the algorithmic structure of \rev{GNN} training and provides adequate  memory capacity/bandwidth,  but still suffers from resource under-utilization and load-imbalance problems due to the irregularity of graphs. Therefore, 
we  propose several optimization strategies concerning  data reuse, data mapping,  graph partition, and dataflow scheduling, etc., to improve the training throughput further. 
To summarize, we make the following main contributions:
\begin{itemize}[leftmargin=1em]

\item We characterize the full-batch \rev{GNN} training and
abstract the involved operations as memory-intensive \texttt{Reduce} and computation-intensive \texttt{Update} operations. 
    (Section \ref{sec:heterogeneous})
    \item We propose \texttt{\rev{GNNear}} accelerator, which leverages both DIMM-based Near-Memory Engines (NMEs) and a Centralized Acceleration Engine (CAE) to provide sufficient memory capacity and match the heterogeneous nature of \rev{GNN} training.
    (Section \ref{sec:proposal})
    \item We propose several optimization strategies to deal with the resource under-utilization and load-imbalance problems caused by the  irregularity in real-world graphs and  improve \texttt{\rev{GNNear}}'s performance further.
    (Section \ref{sec:optimization})
    \item We conduct comprehensive  experiments  on various \rev{GNN} training tasks to validate the superiority of  \texttt{\rev{GNNear}} accelerator over commercial CPU/GPU platforms. 
    (Section 
    \ref{sec:evaluation})

\end{itemize}


%% file: figtex/fig_trend.tex
\begin{figure} [t]
    \centering
    \includegraphics[width=1.0\linewidth]{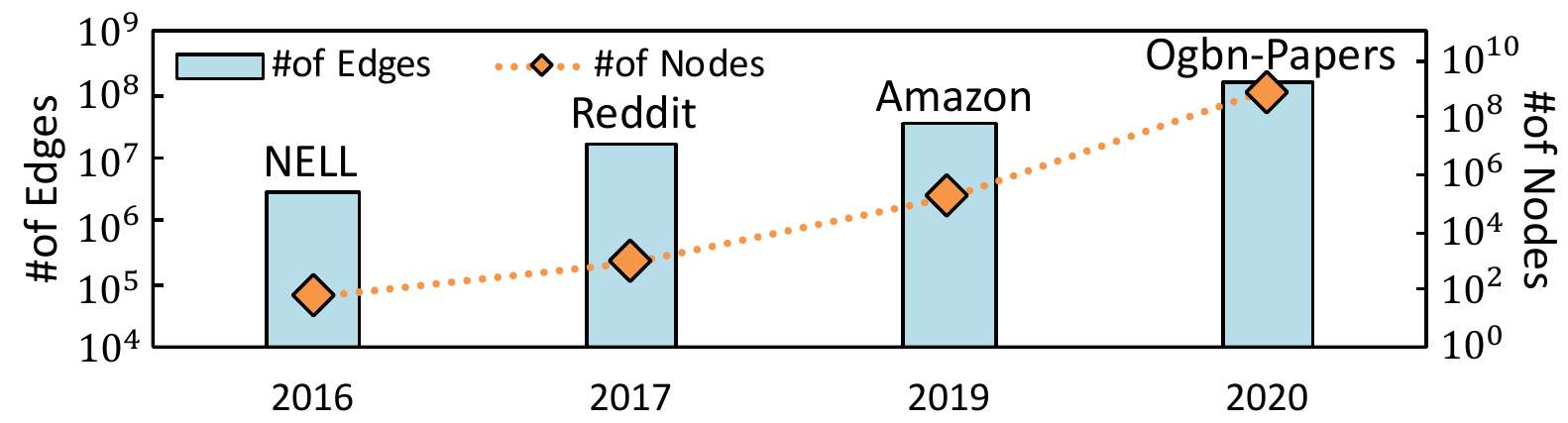} 
    \caption{The trend of increasing graph scale.}
         \label{fig:trend}
\end{figure}

%% file: tex/background.tex
\input{tables/gcn_variants}
\input{figtex/fig_gcn.tex}

\section{Background}
\label{sec:background}
   

\noindent \rev{In this section we introduce the basics of GNN and GNN training, notations are listed in Table~\ref{tab:notations}.}

\subsection{\rev{GNN} Algorithm Basics}

\noindent Generally, a \rev{GNN} model is composed of several \rev{GNN} layers.  Given an input graph $G=(V,E)$, where $V$ and $E$ are sets of vertices and edges, each \rev{GNN} layer computes with two-step operations illustrated in  Figure~\ref{fig:gcn}. The operations are also abstracted as follows:
\begin{equation}
\setlength{\abovedisplayskip}{5pt}
\setlength{\belowdisplayskip}{5pt}
\label{eq:aggregate}
    a_v^{l}=\textsc{Aggregate}(h_u^{l-1}|u\in \widetilde{\mathcal{N}}(v))
\end{equation}
\begin{equation}
\myfont
\setlength{\abovedisplayskip}{5pt}
\setlength{\belowdisplayskip}{5pt}
\label{eq:combine}
    h_v^{l}=\textsc{Combine}(a_v^{l})
\end{equation}
At the $l$-th layer, the \emph{Aggregation} step gathers each vertex $v$'s neighbor features (denoted as $h_u^{l-1}$) and uses an aggregator  to merge the features.  Then the aggregation  result $a_v^l$  is processed with the \emph{Combination} step, which transforms $a_v^l$ to $h_v^l$ using a neural network. After combination, $h_v^l$ serves as the input of layer $l+1$. The final layer's outputs $h_v^L$ will be used as vertex-level representations for various downstream tasks.

The GCN algorithm~\cite{semigcn} adopts \emph{Weighted Sum} as its aggregator and  combines with  a  fully-connected layer:
\begin{equation}
\myfont
\label{eq:gcn}
\setlength{\abovedisplayskip}{5pt}
\setlength{\belowdisplayskip}{5pt}
    a_v^{l}= {\sum_{u\in \widetilde{\mathcal{N}}(v)}{\frac{1}{\sqrt{D_v*D_u}}} h_u^{l-1}},~~h_v^l = ReLU(a_{v}^{l}\cdot W^{l})
\end{equation}
In the formula, $D_v$ and $D_u$ denote the degrees of vertex $v$ and each neighbor $u$.  The neighbor features are aggregated using summation with degree-based normalization.
Apart from GCN, several \rev{GNN} variants are also proposed. As listed in Table~\ref{tab:gnn}, 
GIN~\cite{gin} and SAGEConv~\cite{graphSAGE}
 use \emph{Sum} and \emph{Mean} operators for aggregation.
GAT\cite{wang2019kgat} leverages a self-attention mechanism to implement its aggregator. It first calculates the attention coefficient $\alpha_{uv}$, which measures the importance of vertex $v$ to $v$'s neighbor $u$.
Then the aggregation results are the weighted sum of neighbors' features. Except for GIN, which adopts Multi-Layer-Perception (MLP) for combination, both SAGEConv and GAT use a single fully-connected layer for combination. 
Without loss of generality, we use GCN as a representative example in the following discussions.
\label{sec:gcn_training}

\input{tables/anotations}

\subsection{\rev{GNN} Training}

\noindent To learn useful information from input graphs, a \rev{GNN} model first undergoes the training procedure, which can be divided into the forward pass and the backward pass. As illustrated in Figure~\ref{fig:simple_flow}, 
after a sequence of aggregation and combination operations, a loss function calculates  loss $\mathcal{L}$ with the forward outputs and labels. In the backward pass, each $\nabla \textit{CB}$ step computes the gradients of weights $W^l$ and hidden features $h_v^l$, while  $\nabla \textit{AG}$ aggregates the feature gradients for each vertex along edges.  We can easily derive the formulation of $\nabla \textit{AG}$ through the chain-rule:
\begin{equation}
\setlength{\abovedisplayskip}{4pt}
\setlength{\belowdisplayskip}{4pt}
\begin{aligned}
  \nabla \textit{AG}:~~  {\delta}_v^{l'}  =\texttt{mask}(\sum_{u\in \widetilde{\mathcal{N}}(v)}\frac{1}{\sqrt{D_u*D_v}}{\delta}_u^{l+1})
\end{aligned}
\end{equation}
Where $\delta_u^{l+1}$ denotes the feature gradients of vertex $u$ at layer $l+1$, namely $\frac{\partial \mathcal{L}}{\partial h_u^{l+1}}$.
$\texttt{mask}()$ corresponds to gradients of activation function like \texttt{ReLU}$()$.  Then  $\nabla \textit{CB}$ computes the  weight gradients with the feature gradients of each vertex $v$:
\begin{equation}
\setlength{\abovedisplayskip}{5pt}
\setlength{\belowdisplayskip}{5pt}
\begin{aligned}
 \nabla \textit{CB}: ~~\delta_v^l = \delta_v^{l'}\cdot W^{l+1^T},  ~~  \frac{\partial \mathcal{L}}{\partial W^l} = \frac{\partial \mathcal{L}}{\partial W^l} + a_v^{l^T}\cdot {\delta}_v^l    \end{aligned}
 \label{eq:delta_cb}
\end{equation}
The weight gradients $\frac{\partial \mathcal{L}}{ \partial W^l}$ are initialized to zero at the beginning of each training epoch.  The final weight gradients are used  to update the model weights through  gradient descent: $W^l = W^l -  \eta\frac{\partial \mathcal{L}}{ \partial W^l}$, where $\eta$ denotes the learning rate. The training process iterates for several epochs until  convergence. 

The main computation in both forward and backward passes can be categorized into two different types according to their computation patterns, namely  \texttt{Reduce} and \texttt{Update}. The \texttt{Reduce} operations aggregate features/gradients along the edges of each destination vertex, which are abstracted as: 
\begin{equation}
\setlength{\abovedisplayskip}{4pt}
\setlength{\belowdisplayskip}{4pt}
\label{eq:reduce}
\texttt{Reduce}: ~~Y_v=\sum_{u\in \widetilde{\mathcal{N}}(v)} edge\_w(u,v) \cdot X_u
\end{equation}
Vectors $X_u$ and $Y_v$ can be either hidden features (forward) or feature gradients (backward), while scalar $edge\_w(u,v)$ denotes the edge weight concerning  source vertex $u$ and  destination vertex $v$ (e.g., $\frac{1}{\sqrt{D_u*D_v}}$). 
Each vertex's \texttt{Reduce} operation   accumulates the weighted feature/gradient vectors from all its neighbors. Thus the total number of weighted vector additions for each layer is  $2\times\left|E\right|$.
 
The \texttt{Update} operations perform vector-matrix or vector-vector multiplications to generate new features/gradients for vertices or weights, which are formulated as:
\begin{equation}
\setlength{\abovedisplayskip}{4pt}
\setlength{\belowdisplayskip}{4pt}
    \texttt{Update}:~~ Y_v = X_v\cdot W~~ or~~ Y_v = X_v \cdot Z_v
\end{equation}
Where $W$ denotes the weight matrix (or transposed weight matrix) and $X_v, Z_v$ are feature/gradient vectors (e.g., $X_v = a_v^l$ or $a_v^{l^T}$, and $Z_v = {\delta}_v^l $).   For each layer, {the total number of   vector-matrix or vector-vector multiplications is proportional to $|V|$}. 
In Section~\ref{sec:hetero} we will also demonstrate that \texttt{Reduce} and \texttt{Update} operations have different arithmetic intensity.
\input{figtex/fig_simpleflow.tex}

\subsection{Full-batch VS. Mini-batch Training:} 
\noindent On real-world tasks, the input graphs can be too large to train on a single GPU.
Mini-batch training is then proposed  to mitigate this problem via neighborhood sampling~\cite{fastgcn,clustergcn,graphSAGE,zeng2019graphsaint}. They sample vertices and their neighbors to create a mini batch that can fit into GPUs. However,  it has been widely admitted that due to approximation errors, in some cases, mini-batch training achieves lower accuracy compared to full-batch training ~\cite{graphSAGE,distgnn,roc,tripathy2020reducing,dgcl}. \rev{Therefore, many mini-batch training algorithms focus on proposing accurate sampling methods to improve the model accuracy~\cite{9601152}}. \rev{For full-batching training, how to improve the training efficiency is the key problem~\cite{roc,distgnn,tripathy2020reducing,dgcl}.}
\rev{Considering that} a system supporting full-batch training can also conduct mini-batch training by adding an extra sampler. 
\rev{We focus on full-batch training in this paper.}

%% file: tables/gcn_variants.tex
\newcommand{\tabincell}[2]{\begin{tabular}{@{}#1@{}}#2\end{tabular}}  
\begin{table}[t]
\setlength{\abovecaptionskip}{0cm} 
\caption{Variants of \rev{GNN}s}
\label{tab:gnn}
\resizebox{0.48\textwidth}{!}{
\setlength{\tabcolsep}{1mm}{
\renewcommand{\arraystretch}{1.5}{
\begin{tabular}{c|c|c}
 \toprule
Variant & Aggregation & Combination \\ \midrule 
 GCN~\cite{semigcn}       & $a_v^{l}= {\sum_{u\in \widetilde{\mathcal{N}}(v)}{\frac{1}{\sqrt{D_v*D_u}}} h_u^{l-1}}$&           $Relu(a_{v}^{l}\cdot W^{l})$            \\ \midrule
GIN~\cite{gin}  &           $a_v^{l}=(1+\epsilon)\cdot h_v^l + \sum_{u\in \mathcal{N}(v)}h_u^{l}$              &      \textit{MLP}$(a_v^l,W^l,b^l)$                 \\ \midrule
SAGEConv~\cite{graphSAGE}   &     
 $ {a}^{l}_v = 
        {Mean}(h_u^l|u\in \widetilde{\mathcal{N}}(v))$
              &      $Relu(a_{v}^l \cdot W^{l})$                 \\ \midrule
GAT~\cite{wang2019kgat}       & 
\tabincell{c}{$    \alpha_{vu}=Softmax(\sigma(\mathbf{a}^T[W_1^lh_v^l||W_2^lh_u^l] ))
$\\$    a_v^l = \sum_{u\in \widetilde{\mathcal{N}}} \alpha_{vu}\cdot h_u^l
$ }
              &       $Elu( a^{l}_v \cdot W^l)$               \\ \bottomrule 
\end{tabular}}}}
\end{table}

%% file: figtex/fig_gcn.tex
\begin{figure} [t]
    \centering
    \includegraphics[width=0.98\linewidth]{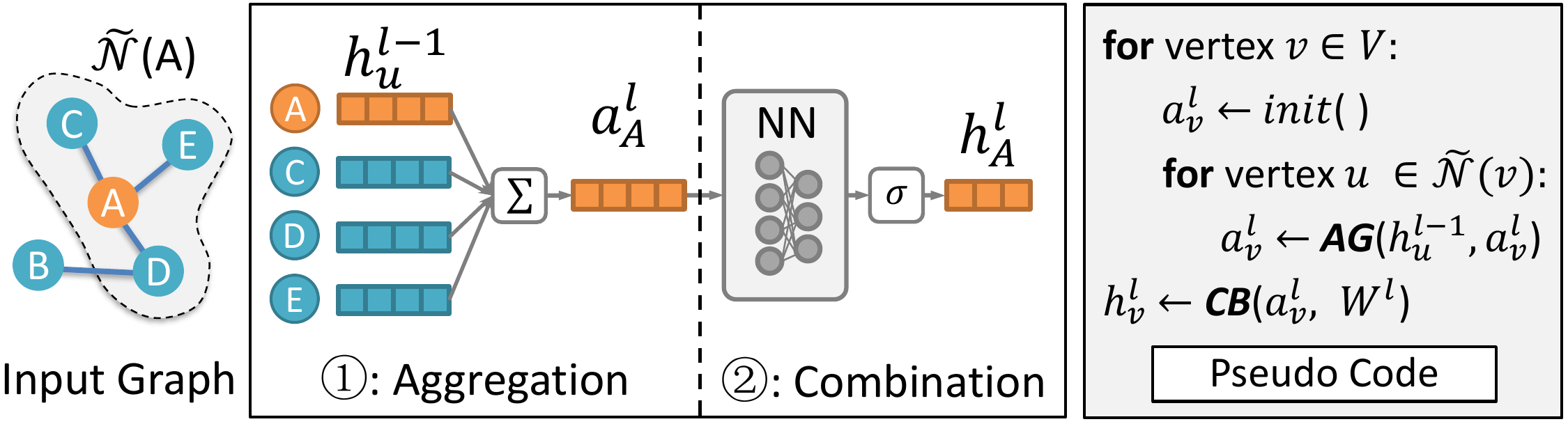} 
    \caption{The GCN inference workflow.
    }
         \label{fig:gcn}
\end{figure}

%% file: tables/anotations.tex
\begin{table}[t]
\caption{\rev{List of Notations Used in \rev{GNN} Training Algorithms}}
\label{tab:notations}
\centering
\resizebox{0.48\textwidth}{!}{
\renewcommand{\arraystretch}{1}{
\begin{tabular}{c|l}
\toprule
Notation   & Description            
\\ \midrule
$V, E$ & Sets of vertexes and edges \\
$\mathcal{N}(v)$
 & The neighbors of vertex $v$  \\ 
 $\widetilde{\mathcal{N}}(v)$
 & The set containing $v$ and $v$'s neighbors: $\{\mathcal{N}(v)\}\cup\{v\}$
 \\
$D_v$    & The degree of vertex $v$                     \\ \midrule
$h_v^l$  & Hidden feature vector of vertex $v$ at the $l$-th layer \\
$a_v^l$ & Aggregated feature vector of vertex $v$ at the $l$-th layer \\
$W^l$    & Weight matrix in layer $l$                   \\ \midrule
$\mathcal{L}$ & The loss value calculated with outputs and  labels\\
$\delta_v^l$ &  The gradients of $h_v^l$, namely  $\frac{\partial \mathcal{L}}{\partial h_v^l} $                                               \\ 
${\delta}{_v^l }'$ & Aggregated masked gradients for vertex $v$ \\
\bottomrule
\end{tabular}}}
\end{table}

%% file: figtex/fig_simpleflow.tex
\begin{figure} [t]
    \centering
    \includegraphics[width=0.98\linewidth]{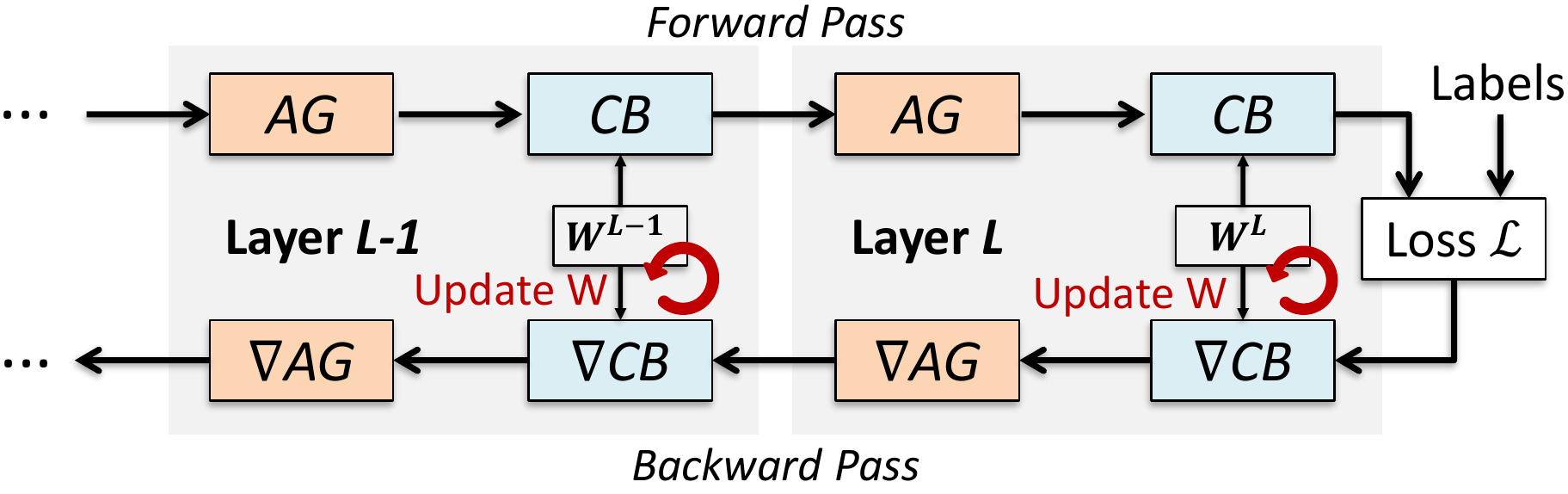}
    \vspace{-2pt} 
    \caption{GCN training dataflow. 
    }
         \label{fig:simple_flow}
\end{figure}

%% file: tex/motivation.tex
\section{Challenges Analysis}
\label{sec:heterogeneous}
\subsection{Characterising  \rev{GNN} Training}
\noindent\textbf{Large Memory Footprint:}
The ever-increasing graph scale shown in Figure~\ref{fig:trend} poses a great challenge to full-batch \rev{GNN} training. For example, training on the middle-scale Reddit dataset~\cite{graphSAGE} containing 114 million edges requires 58 GB of memory to hold all the intermediate data and incurs over 300 GB of DRAM access~\cite{fuseGNN} with PyG framework~\cite{pyg}. DGL framework ~\cite{DGL} has an optimized memory management but is still hard to train on a larger dataset like Amazon~\cite{clustergcn} with a single GPU. 
According to the analysis~\cite{roc},  we need to buffer much temporal data during training. 
The Ogbn-Papers dataset containing 111 million vertices~\cite{hu2020ogb} consumes at least 568 GB of  memory to train a two-layer model with a hidden size of 256. What is worse, the data gathering operations along edges incur $O(L\times|E|\times d)$ memory accesses ($L$ denotes the number of  layers while $d$ is the hidden feature dimension), making the system also memory-bandwidth-bounded. 

\noindent\textbf{Training Bottlenecks:} We profile the execution time of \rev{GNN}-training operators using PyTorch-profiler~\cite{profiler}  on an NVIDIA V100-32GB GPU with PyG~\cite{pyg} framework.  Four \rev{GNN} algorithms introduced in Section~\ref{sec:background} and four middle-scale graph datasets, namely  PubMed (PB)~\cite{bojchevski2017deep}, Flickr (FL)~\cite{zeng2019graphsaint}, Amazon-Computer (AC)~\cite{shchur2018pitfalls}, and Reddit (RD)~\cite{graphSAGE},  are adopted as benchmarks. According to Section~\ref{sec:gcn_training},  
 we have classified the main \rev{GNN} training operations as  \texttt{Reduce} and  \texttt{Update} operations to process feature/gradients aggregation and combination.  The remaining operations, such as loss computation, are classified as \texttt{Others}.
 As shown in Figure~\ref{fig:profiling}, in general, \texttt{Reduce} and \texttt{Update} operations are the main bottlenecks during \rev{GNN} training.  The \texttt{Reduce} operations are the most time-consuming on V100 GPU in most cases,  but \texttt{Update} operations also take considerable time on PB and FL datasets, especially for GAT. On CPU platforms, \texttt{Update} operations will take up more time due to CPUs' poor computation capacity.
 \input{figtex/fig_training_time_breakdown}

\input{tables/profiling_on_cpu}

 \input{figtex/arc_overview}

\noindent\textbf{The Heterogeneous Nature:} 
\label{sec:hetero}
\rev{GNN}s' weights are usually small in size. For instance, a 4-layer GCN with input-size, output-size, and  hidden-size of 256 only has about 1MB of weights. It is feasible to store weights to on-chip SRAM and reuse them among vertices.
Therefore, assuming $d_{in}=d_{out}=d$ for simplicity, where $d_{in},d_{out}$ represent the input and output dimensions of a layer, the theoretical arithmetic intensity of \texttt{Update} with the form of $Y_v = X_v \cdot W$ is $\frac{2\times d_{in}\times d_{out}}{4B\times (d_{in}+d_{out})}$ = $0.25\times d$ Ops/Byte (FP32).
For \texttt{Update} operations with the form of $Y_v = X_v \cdot Z_v$ ($X_v = a_v^{l^T}$ and $Z_v = {\delta}_v^l $ in Equation.~\ref{eq:delta_cb}), the shape transformation is  $(d \times 1)\cdot (1\times d) = (d\times d)$. We have to read vectors $X_v, Z_v$, but do not need to write back the weight gradients $Y_v$.  Thus, the arithmetic intensity is $\frac{d\times d }{4B\times 2d} = 0.125\times d$~Ops/Byte.
On the contrary, for \texttt{Reduce} operations, massive amounts of features/gradients  should be loaded from DRAM and can hardly be reused. Then the arithmetic intensity is $\frac{2\times|\widetilde{\mathcal{N}}(v)|\times d_{in}}{4B\times (|\widetilde{\mathcal{N}}(v)| \times d_{in} + d_{out})} \approx 0.5$ Ops/Byte, if using \emph{Weighted Sum} as the aggregator.  
Obviously, \texttt{Reduce} operations are much more memory-intensive than \texttt{Update}. 
We also conduct CPU-based real-system profiling using Intel-Vtune~\cite{Vtune}. As shown in Table~\ref{tab:character},
\texttt{Reduce} operations show a much lower arithmetic intensity and worse data locality than \texttt{Update} and \texttt{Others}, which is inefficient to accelerate with traditional neural network accelerators~\cite{eyeriss, chen2014diannao, dadiannao, ten_lessons, zhang2015optimizing} optimized for computation-intensive workloads.



\subsection{Near-Memory-Processing to the Rescue?}


\noindent Recently, Near-Memory-Processing (NMP) paradigm   has  been proposed to provide memory-capacity proportional bandwidth and computation capacity.  Chameleon~\cite{chameleon} is a pioneering  work that integrates CGRA cores  to  buffer-chips of  DDR4 LRDIMMs~\cite{LRDIMM} to enable general-purpose near-memory computation.  RecNMP~\cite{recnmp} and TensorDIMM~\cite{tensordimm} also propose to accelerate  Deep Learning Recommendation Models (DLRMs) adopting such a paradigm.
These works motivate us to leverage the DIMM-based NMP technique to accelerate full-batch \rev{GNN} training.  

However,  designing a NMP accelerator suitable for full-batch \rev{GNN} training is still a challenging task. Firstly, unlike DLRMs, which only need to consider the memory-intensive  embedding table gathering operations \cite{facebookrec}, \rev{GNN} training also has computation-bounded \texttt{Update} operations due to its heterogeneous nature. For instance, a two-layer GIN model (a \rev{GNN} variant in Table~\ref{tab:gnn}) trained on Ogbn-Papers dataset with a hidden  size of 256 incurs more than \ogbnUpdateComputation~computation for just one iteration. It takes a  server CPU several minutes to finish. 
Secondly, unlike DLRMs that respond to random quires,  \emph{\rev{GNN} training must consider the graph structure}. Due to the irregularity of real-world graphs,  a na\"ive NMP-based \rev{GNN} training system will have severe resource under-utilization and load-imbalance problems.

%% file: figtex/fig_training_time_breakdown.tex
\begin{figure} [t]
    \centering
    \includegraphics[width=1.0\linewidth]{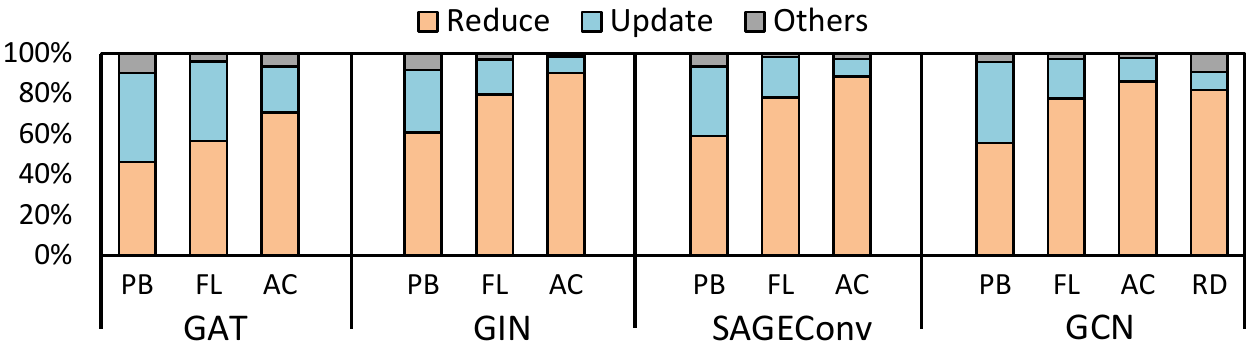} 
    \caption{Training time breakdown on V100 GPU.}
         \label{fig:profiling}
\end{figure}

%% file: tables/profiling_on_cpu.tex
\begin{table}[t]
\caption{Characterization of GCN training on CPU}
\label{tab:character}
\centering
\setlength{\tabcolsep}{5.0mm}{
\renewcommand{\arraystretch}{0.85}{
\resizebox{0.48\textwidth}{!}
{\centering
\begin{tabular}{c|c|c|c}
\toprule
                Operations         & {Reduce}  & {Update} & {Others} \\ \midrule
Ops per DRAM Byte &      0.84             &  22.6    &  715     \\ 
L2 Cache MPKI            & 65.7                   & 3.4    &  1.2 
      \\ 
L3 Cache MPKI            &          15.2          &    0.33  &  0.01      \\ \bottomrule
\end{tabular}}}}
\end{table}

%% file: figtex/arc_overview.tex
\begin{figure*} [t]
    \centering
    \includegraphics[width=0.98\linewidth]{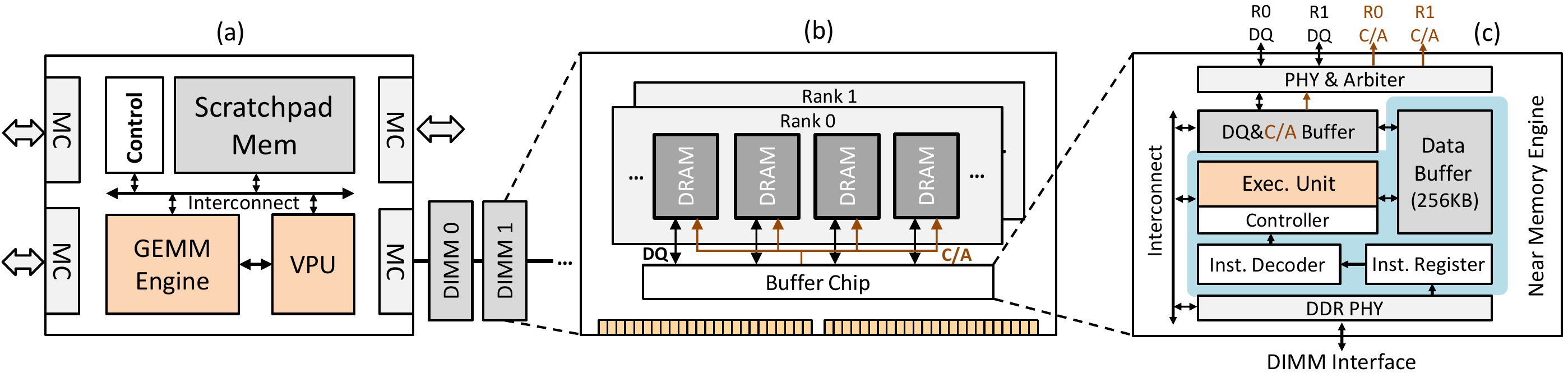} 
    \caption{ \texttt{\rev{GNNear}} overview. (a): Centralized Acceleration Engine (CAE). (b): NMP-enabled  DIMM.  (c): Near-Memory Engine (NME).}
         \label{fig:arc}
         \vspace{-1em}
\end{figure*}

%% file: tex/architecture.tex
\section{\textsc{\rev{GNN}ear} Architecture}

\input{figtex/fig_base_workflow}

\noindent To overcome the mentioned challenges, we propose \texttt{\rev{GNN}ear}, a hybrid \rev{GNN} training accelerator combining both DIMM-based near-memory processing engines and a centralized acceleration engine. It matches the heterogeneous nature of \rev{GNN} training and provides sufficient memory capacity for full-batch \rev{GNN} training.

\label{sec:proposal}
\label{sec:accelerator}
\subsection{Overview}
\noindent\textbf{Design:}
As illustrated in Figure~\ref{fig:arc}, \texttt{\rev{GNN}ear} accelerator consists of a Centralized Acceleration Engine (CAE) and multiple NMP-enabled DIMMs. The CAE resembles Google's TPU, which equips a powerful GEMM engine and a vector-processing unit (VPU) 
to deal with  computation-intensive \texttt{Update} operations. 
NMP-enabled DIMMs are connected to CAE's four memory channels, each containing a Near-Memory Engine (NME) for memory-intensive \texttt{Reduce} operations. 
Such a CAE/NMEs hybrid architecture matches the heterogeneous nature of \rev{GNN} training. More importantly, we can scale up the  capacity, processing ability, and aggregated memory bandwidth by connecting more DIMMs to the memory channels.

\noindent\textbf{Base Workflow:} Figure \ref{fig:workflow} depicts the base  workflow of \texttt{\rev{GNN}ear}.  
The input features of graph $G$ in Figure \ref{fig:workflow}-(a) are initially stored in DRAM. 
We partition $G$'s vertices evenly and assign them to different DIMMs. In the figure, we use two DIMMs (DIMM-0 and DIMM-1) for illustration,  which hold the data of blue  and grey vertices, respectively. According to the \rev{GNN} training flow introduced in Section~\ref{sec:background}, we traverse every destination vertex (e.g.,  $v_5$ in Figure \ref{fig:workflow}-(b)) in each forward and backward step and conduct \texttt{Reduce} or \texttt{Update} operations. For \texttt{Reduce}, CAE sends customized instructions through the memory  interfaces to NMEs. NMEs decode the instructions and perform partial reduction with the assigned source vertices locally (e.g., DIMM-0 and DIMM-1 compute feature vectors $a_0'$ and $a_1'$, respectively). Then, the partial results are read out by CAE (operation \textcircled{\scriptsize {1}} in the figure). CAE merges these partial results (i.e., $a_0'+a_1'$) to get the final reduction results. For \texttt{Update}, CAE computes with the merged results buffered on-chip or directly reads the required data  without near-memory reduction. The updated features or gradients will be written back to DRAM if necessary (operation \textcircled{\scriptsize {2}}).  After a training epoch, CAE updates the model weights with accumulated weight gradients. As Figure~\ref{fig:workflow}-(c) shows, in each epoch we process \texttt{Reduce} and \texttt{Update} operations of the adjacent destination vertices \emph{in a pipelined manner} since they have no data dependency. 
Such a training process executes for many epochs until convergence.

With such a near-memory reduction workflow,
the \texttt{Reduce} operations in Equation.~\ref{eq:reduce} are then changed to
\begin{equation}
\setlength{\abovedisplayskip}{2pt}
\setlength{\belowdisplayskip}{2pt}
\myfont
  Y_v^{i'}=\sum_{u\in{\widetilde{\mathcal{N}}(v)}  | u\in \textit{DIMM}_i } edge\_w(u,v)\cdot X_u ,~~~Y_v = \sum_{i=1}^{\#\textit{DIMM}} Y_v^{i'}
  \label{eq:new_reduce}
\end{equation}
The off-chip data read is reduced from the original $|\widetilde{\mathcal{N}}(v)|\times d$ to no more than \emph{\#DIMMs}$ \times d$ for vertex $v$. The latter is usually much smaller than the former.
Considering that NMEs in different DIMMs work in parallel, \texttt{\rev{GNN}ear} can utilize the high aggregated local bandwidth.
Moreover, for computation-intensive \texttt{Update} operations, CAE provides sufficient computation capacity. 

\subsection{CAE Architecture}
\noindent Centralized Acceleration Engine (CAE) is mainly responsible for the computation-intensive \texttt{Update} operations. It also merges partial reduction results $Y_v'$ produced by NMEs. As shown in Figure~\ref{fig:arc}, CAE has both a GEMM engine (implemented with systolic array architecture) and a vector-processing unit (VPU). A scratchpad memory is equipped to buffer the temporary data and model weights. 
Four customized memory controllers support sending \texttt{\rev{GNN}ear} instructions to NMEs. A controller (can be implemented with an OoO RISC-V or ARM core) schedules the whole training process according to the input graph's adjacent matrix and the model's configuration.  A high-bandwidth on-chip network connects all the components.

\subsection{NME Architecture}
\noindent The Near-Memory Engine (NME) resides in the buffer chip of DDR4 LRDIMM~\cite{LRDIMM}. Each NMP-enabled DIMM equips one NME. In Figure~\ref{fig:arc}-(c), we mark the customized components with blue. There are five main components:  Instruction Register, Instruction Decoder, Execution Unit,   Data Buffer, and a Controller. 
NME  receives NMP instructions from CAE and puts them in the Instruction Register. Instruction Decoder decodes each instruction. Then, the light-weighted controller starts local execution following the instruction. The Execution Unit handles data calculation.
The controller is also responsible for generating regular DDR4 Command/Address and data signals (DDR.C/A and DDR.DQ) and sending them to all DRAM devices across  parallel ranks in a DIMM (two ranks in the example).  Since NMEs can access their local DRAM devices in parallel, the aggregated local bandwidth is much higher. 
Furthermore, we equip an SRAM data buffer in NME to  explore data locality in graph structure~(Section \ref{sec:reuse}).
Apart from near-memory processing, if NME receives standard DDR commands from the CAE-side memory controller, it will bypass execution units and directly conducts  Read/Write/Precharge commands, etc.  
\input{figtex/fig_execution_unit}

The Execution Unit (EU for short)  in each NME  is responsible for near-memory partial reduction computation of \texttt{Reduce}  operations. According to Equation~\ref{eq:new_reduce}, partial reduction  is formulated as the weighted sum of $n$ feature/gradient vectors: $Y_v' = \sum_{u=1}^n edge\_w(u,v) * X_u$, where $edge\_w$ denotes edge weight (scalar). For GCN, the edge weight is $\frac{1}{\sqrt{D_uD_v}}$. For GAT, it is an estimated importance factor. As shown in Figure \ref{fig:eu}, The EU adopts an intra-feature parallelism data flow. There are in total $m$ PEs, each computing eight elements every cycle. The results are added to partial sums stored in registers.  Since each element in a vector $X_u$ shares the same edge weight, $edge\_w$ is broadcast to all multipliers. Therefore, $m$ PEs compute $8 \times m $ elements of a vector in parallel. 
If $X_u$ is longer than $8\times m$, it needs multiple rounds to finish the computation. EU also supports \emph{Sum} and \emph{Mean}  operators used in GIN and SAGEConv by setting $edge\_w$ to 1 and $1/n$, respectively.

\subsection{Data Reuse with Narrow-Shard Strategy}
\label{sec:reuse}

\noindent {To save DRAM access further, we explore data reuse  through graph sharding strategies}~\cite{chi2016nxgraph,  hygcn,stevens2021gnnerator,dai2018graphh}. We tune the sharding parameters to make it suitable for NMEs.
In Figure~\ref{fig:cache}, DIMM-0 computes partial reduction of destination vertices $v_1$-$v_8$. We partition the edges into multiple $R\times C$ shards and process one shard each step. For each shard, we first load $R$ source vertices from DRAM  (e.g., operation \textcircled{\scriptsize {1}}), then the computation within the same shard (operations \textcircled{\scriptsize {2}},\textcircled{\scriptsize{3}}) can reuse the loaded source vertex data (e.g., $X_4$). {The partial results of $C$ destination vertices ($Y_5',Y_6'$) are also reused by the following shards of the same column. Therefore, we have to buffer $R$ source vertices and $C$ partial sum on-chip, requiring at least $(R+C)\times d$ space. 
}
{We assume that at most $128$ data vectors can be buffered on-chip and explore different shard configurations. As shown in Figure~\ref{fig:rc}, setting $R=1$ and $C=127$ brings the lowest DRAM access on both Amazon~\cite{clustergcn} and Reddit~\cite{graphSAGE} datasets. \input{figtex/fig_shard_reuse}
\input{figtex/fig_rc_exploration}
It is easy to explain such a result: The $R$ dimension affects inter-shard source vertex reuse. For example, in Figure \ref{fig:cache}, if $R=4$ then source vertex data $X_1$-$X_4$ are always on-chip from step 1 to step 11. However, for real graphs with millions of vertices, Even setting a large $R$ (e.g., R = 127) can hardly bring any meaningful inter-shard reuse. The loaded source vertices are evicted quickly due to poor data locality. Therefore, it is natural to set $R=1$ and enlarge $C$ for the best  inner-shard source vertex reuse. We call it \emph{Narrow-Shard Strategy}.}
Moreover, the Narrow-Shard strategy supports skipping empty shards to save useless data read. (e.g., step 8 to step 9 in  Figure~\ref{fig:cache} skips two empty shards and avoids loading $X_2$,$X_3$).

\subsection{Instruction Set Design}
\label{sec:ISA}
\noindent We design  \rev{GNN}ear-ISA to drive the shard-based near-memory reduction workflow. As demonstrated in Figure~\ref{fig:ISA}, the base \rev{GNN}ear-ISA consists of three types of instructions:

\noindent\textbf{L-Type:} L-Type instruction loads vertex data  from DRAM devices to NME's data buffer. According to our Narrow-Shard strategy,  one data vector $X_u$ of vertex $u$ is loaded each time. Therefore, L-type instruction has \texttt{Daddr} and \texttt{Vector\_Size} fields to indicate the  start address and size of vector $X_u$. NME's controller generates standard DDR read commands according to the received L-type instruction and sends them to DRAM devices to load the data. For standard DDR4 with a burst length of 8 (BL-8), loading a vector longer than $64$B  demands multiple burst-read commands.  

\noindent\textbf{C-Type:} C-Type instruction controls the near-memory calculation. The \texttt{Op} field defines the type of operation. The \texttt{{Edge\_W}} field  (we use 16 bits for BF16 format)  carries the scalar edge weight, namely $edge\_w(u,v)$ in Equation~\ref{eq:reduce} ($e_{u-v}$ in short) that will be multiplied to source vertex's data vector $X_u$.  After multiplication, we index the buffered partial results by \texttt{Dst\_Index} (denotes the destination's \emph{relative index} within a shard) and add it to the partial results $Y_v'$. Taking operation \textcircled{\scriptsize{2}} in Figure~\ref{fig:cache} as an example, we multiply edge weight $e_{4-5}$ and vertex $v_4$'s  data $X_4$ and then index $v_5$'s partial results $Y_5'$ with \texttt{Dst\_Index} = 0 ($v_5$ is the first destination vertex within the shard).  
C-Type instructions rely on CAE to analyze the adjacent matrix  and determine which edge should be processed. CAE is also responsible for calculating the edge weights. Such a design  keeps the NME-side controller as light-weighted as possible.

\input{figtex/fig_ISA}

\noindent\textbf{R-Type:} R-Type instruction is used to read out partial results from NMEs' data buffer upon all shards within the same interval (namely a column of shards) finish computing.  The \texttt{Dst\_Index} and \texttt{Vector\_Size} bits jointly determine which bytes to read. A sequence of R-Type instructions will be issued  to read out multiple destination vertices of an interval.

All instructions have \texttt{DIMM} fields indicating which DIMM should receive the instruction. 
Since R-Type and C-Type operations do not involve DRAM data access, their latency is  determined and denoted as $\textit{tNME\_RD}$ and $\textit{tNME\_CD}$, respectively. The L-Type instruction is decoded into standard DDR commands to read data from DRAM devices. Our data mapping  guarantees that the feature bytes of a vertex will always be mapped into the same DRAM row. Thus, the latency is $\textit{tCL}+\lceil\frac{Vector\ Size}{Burst\ Size}\rceil\times \textit{tBL}$ (row buffer hit) or $\textit{tRC}+\textit{tRCD}+\textit{tCL}+\lceil\frac{Vector\ Size}{Burst\ Size}\rceil\times \textit{tBL}$ (row buffer miss). CAE's memory controllers rely on these timing constraints to schedule NMP instructions. More importantly, 
since all the instructions are issued by CAE and then executed by NMEs under determined timing constraints, \emph{the NMEs do not require any explicit synchronization.}

\input{figtex/fig_mapping}

\subsection{Data Mapping}
\label{sec:mapping}
\noindent Figure~\ref{fig:mapping} demonstrates the base data mapping. The intermediate data during \rev{GNN} training can be accessed using indexes composed by [{\texttt{Vertex\_Index}}$|$\texttt{Type}$|$\texttt{Data}]. \texttt{Data} indexes the bytes within a vector.   \texttt{Type}  indicates the data's type, including feature $h_v^l$, gradients ${\delta}_v^{l}$ or  aggregation results $a_v^l$, etc. We split a vector to the parallel ranks of each DIMM for higher local bandwidth. Each sub-vector is stored sequentially in a DRAM row.  Consecutive vertices are also stored in the same row. Thus, the adjacent shards are more likely to read source vertex from the same activated row (open-page policy). 
We store adjacent vertices in different banks for better bank-level parallelism.  Note that the mapping can be adjusted according to different tasks and system configurations. For example, we can assign more column bits to \texttt{Data} field to hold longer  feature vectors.

%% file: figtex/fig_base_workflow.tex
\begin{figure} [t]
    \centering
    \includegraphics[width=1.0\linewidth]{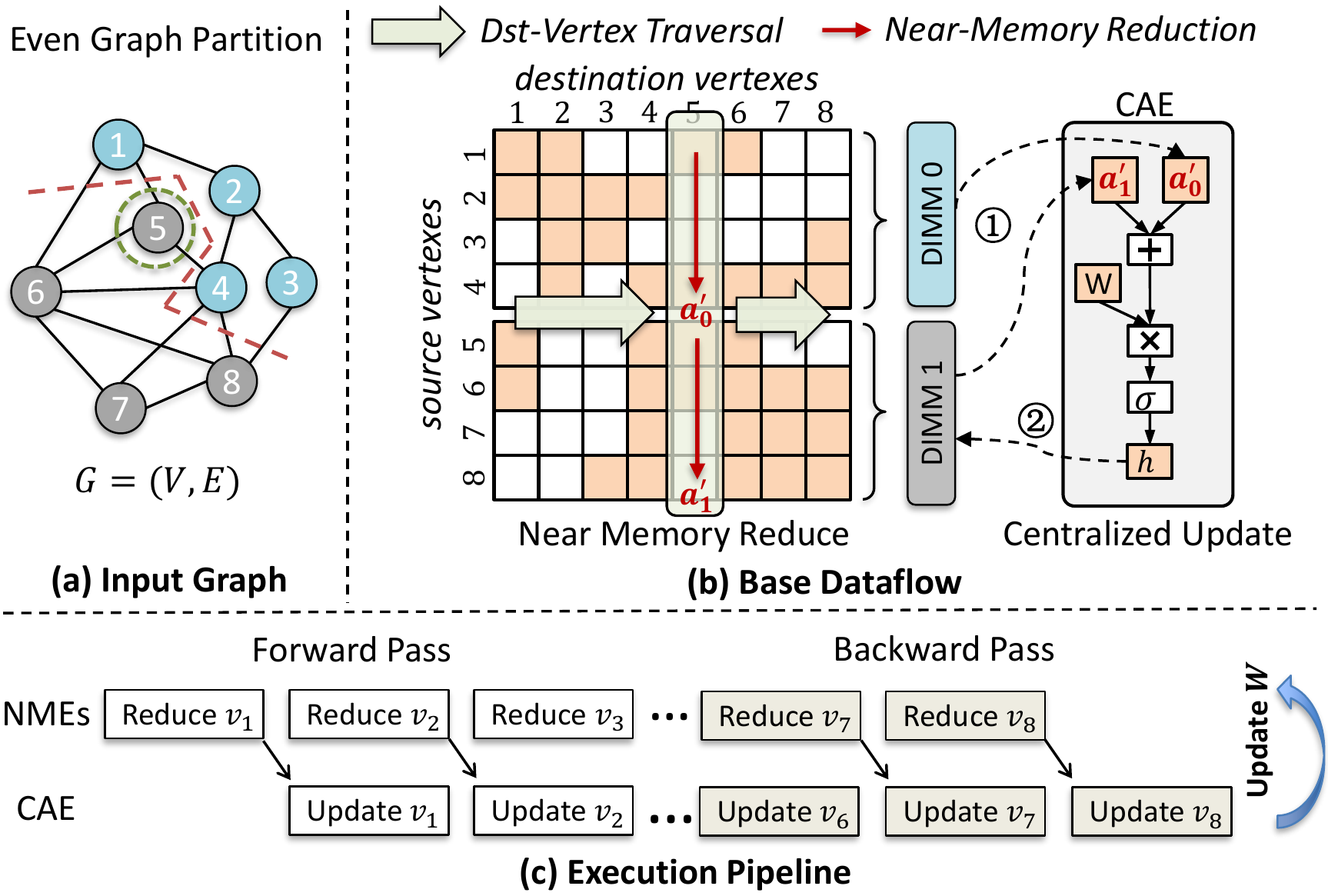} 
    \caption{Base workflow of \texttt{\rev{GNN}ear}. }
         \label{fig:workflow}
\end{figure}

%% file: figtex/fig_execution_unit.tex
\begin{figure} [t]
    \centering
    \includegraphics[width=1.0\linewidth]{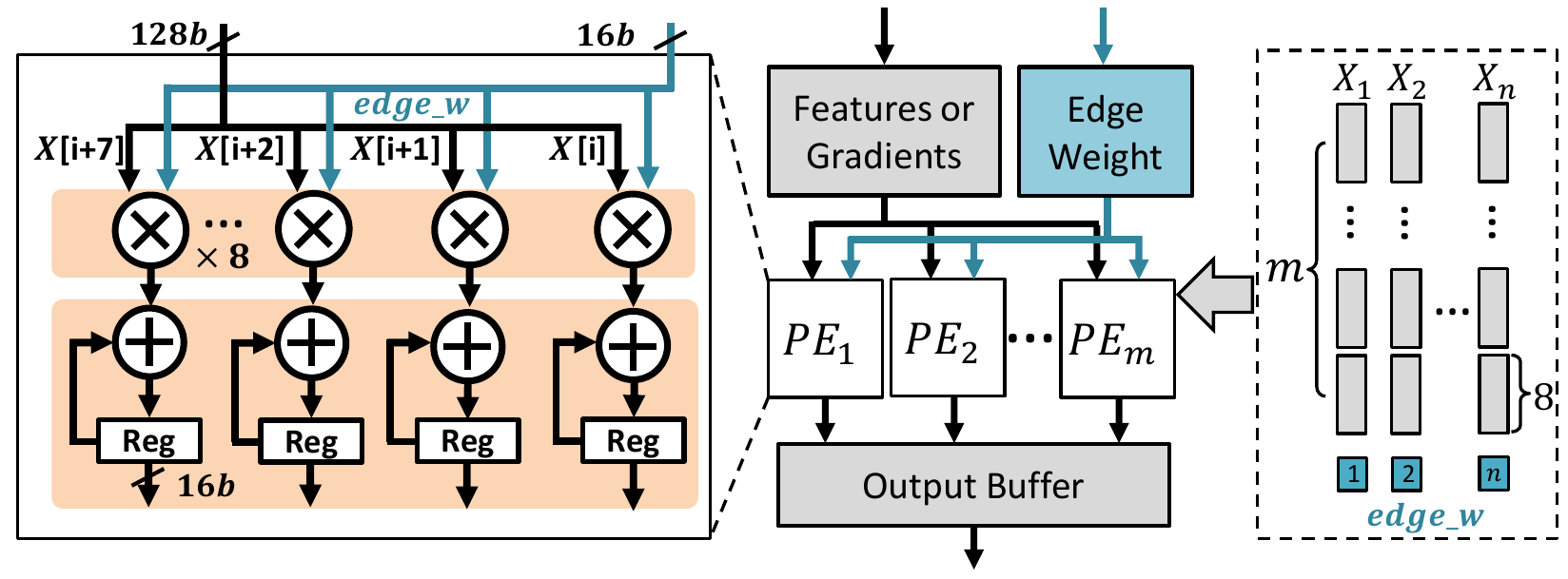} 
    \caption{Execution Unit (EU) architecture in each NME.}
         \label{fig:eu}
\end{figure}

%% file: figtex/fig_shard_reuse.tex
\begin{figure} [t]
    \centering
    \includegraphics[width=0.98\linewidth]{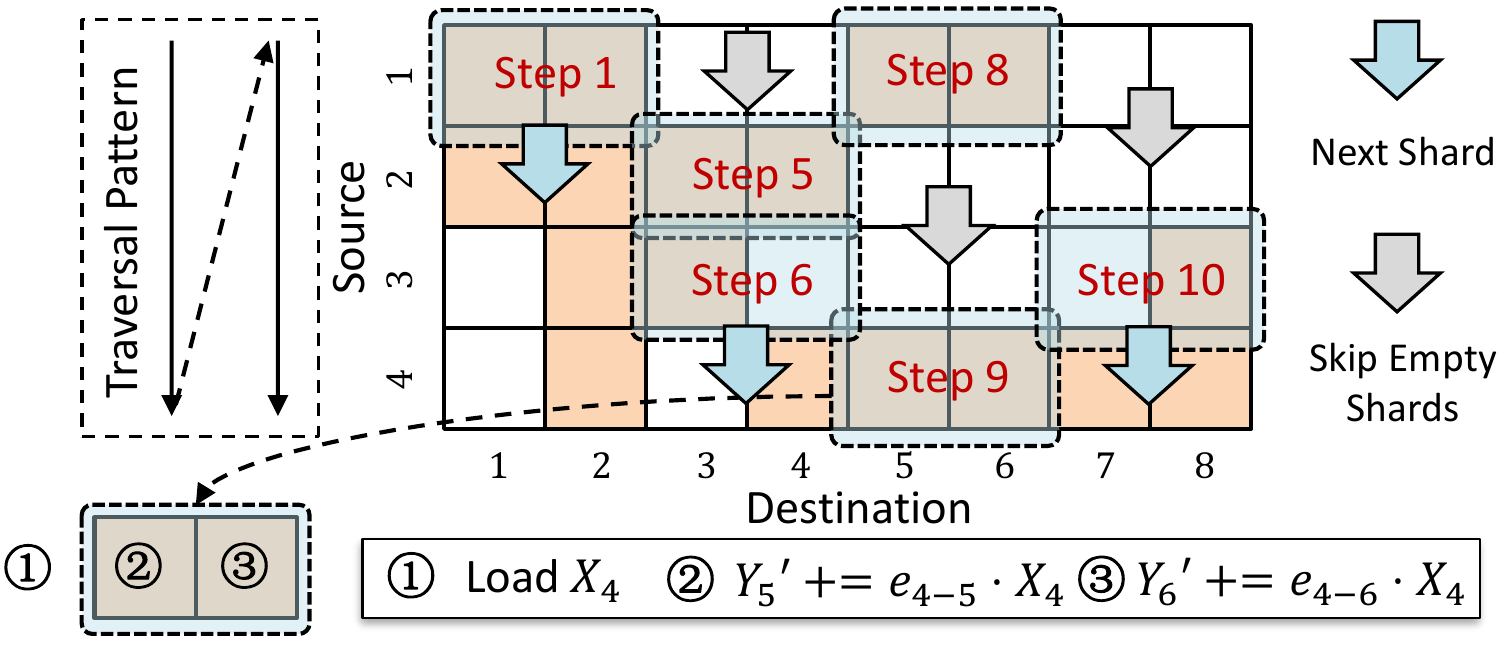} 
    \caption{Narrow-shard-based data reuse.}
         \label{fig:cache}
\end{figure}

%% file: figtex/fig_rc_exploration.tex
\begin{figure} [t]
    \centering
    \includegraphics[width=1.0\linewidth]{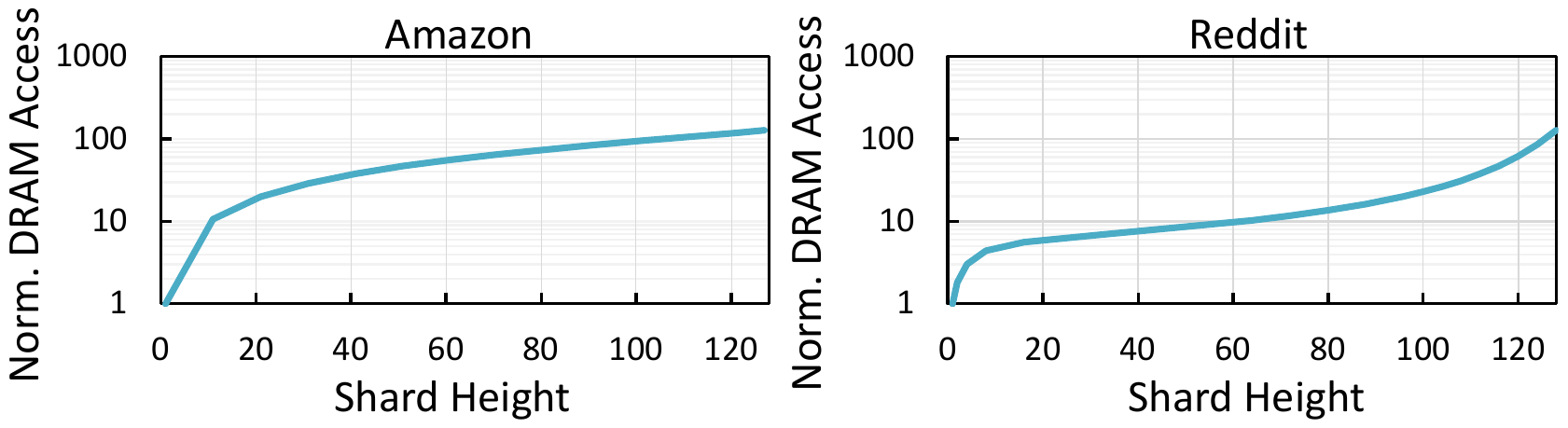} 
    \caption{ Exploration of different shard shapes. }  
         \label{fig:rc}
\end{figure}

%% file: figtex/fig_ISA.tex
\begin{figure} [t]
    \centering
    \includegraphics[width=1.0\linewidth]{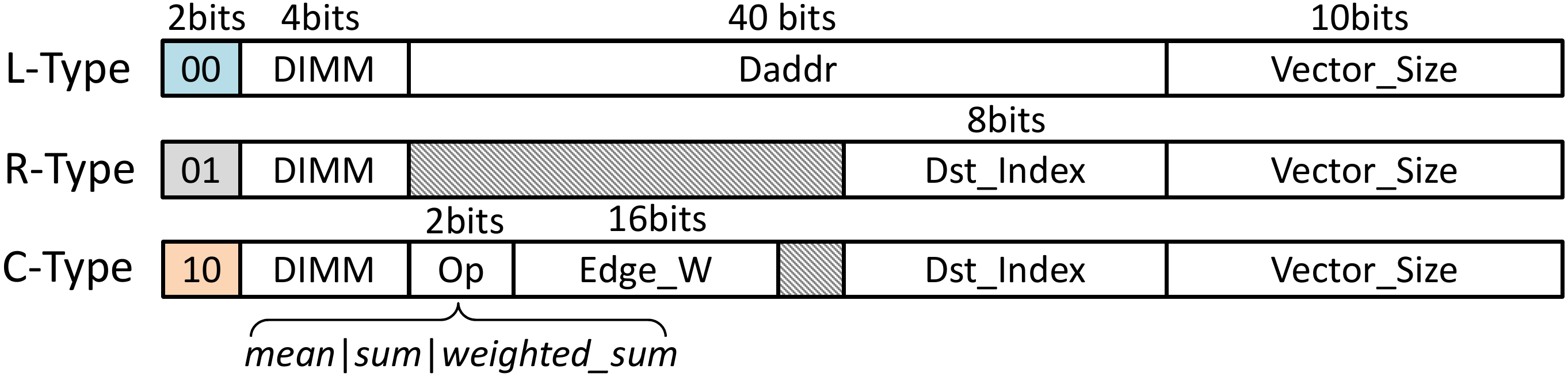} 
    \caption{Base \rev{GNN}ear ISA.}
         \label{fig:ISA}
         \vspace{-0.2em}
\end{figure}

%% file: figtex/fig_mapping.tex
\begin{figure} [t]
    \centering
    \includegraphics[width=0.98\linewidth]{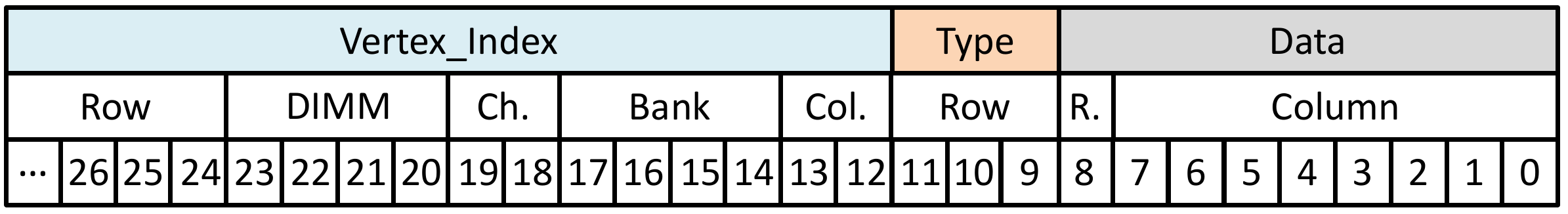} 
    \caption{Data mapping for vertex data. `R.' is short for Rank.}
         \label{fig:mapping}
\end{figure}

%% file: tex/system.tex
\section{Optimizations}
\label{sec:optimization}

\noindent The proposed \texttt{\rev{GNN}ear} architecture and base workflow perfectly match the algorithmic structure of full-batch \rev{GNN} training.
However, it can hardly handle the irregularity of input graphs and will thereby suffer from resource under-utilization and load-imbalance problems.
Therefore, we propose several optimization strategies further to improve \texttt{\rev{GNN}ear}'s performance.

\subsection{Hybrid Graph Partition}

\noindent The base partition strategy (Figure~\ref{fig:workflow}) places source vertices to DIMMs evenly. However, some real-world graphs contain enormous low-degree vertices.
For instance, in Figure~\ref{fig:degree}, about 50\% of vertices in the Amazon graph have degrees $\leq$ 25 (Reddit only has 12\%).  Too many low-degree vertices will cause severe resource under-utilization problems. An extreme case is that vertex $v$ has 16 neighbors placed in 16 different DIMMs. Since each DIMM only contains one neighbor vertex, NMEs cannot perform the partial reduction,
heavily under-utilizing the rich local bandwidth. 
Uneven graph partition considering graph structure information can potentially improve data locality~\cite{gnn_advisor,rubik,clustergcn}. However, they usually require complicated pre-processing and are notoriously  time-consuming on large graphs~\cite{community_gpu}. Therefore, we propose a simple yet practical Hybrid Graph Partition (HGP) strategy. 
According to the power-law hypothesis~\cite{powerlaw}, in real-world graphs, the neighbors of low-degree vertices are probably  the same \emph{super-nodes}. In other words, there are some "dense" rows in a graph's adjacent matrix. Each element in these rows  will incur a partial result readout operation, if these rows are placed in different DIMMs.   
Alternatively, we can put them into the same DIMM  to improve the data locality. However, simply consolidating high-degree vertices will also cause  load-imbalance problems. We choose to duplicate high-degree vertices and   
let each DIMM compute different destinations.

\input{figtex/fig_degree.tex}
\input{figtex/fig_partition.tex}

As shown in Figure~\ref{fig:partition}, all vertices are still evenly partitioned among memory channels. In each channel, we evenly partition the low-degree vertices (Mode-(a)). For high-degree vertices, we duplicate them to all DIMMs and assign the computation tasks to distinct DIMMs (Mode-(b)).     The red circles denote the saved vertex read (we only mark one column).  Such a hybrid partition strategy reduces the data read via channel-0 from 26 to 15  in the example.  
We use an adjustable parameter $\lambda$ to control the duplication ratio and ensure it will not exceed the DIMMs' capacity. HGP  can be executed offline before deploying a training task to \texttt{\rev{GNN}ear}.


\noindent\textbf{Hardware Support:} After adopting the HGP strategy,  we have to update all the duplicated vertices, incurring more off-chip data write operations via the low-bandwidth memory channels.  To tackle this challenge, we propose to update duplicated vertices using \texttt{broadcast-write} operations.
It is feasible to support broadcast-write in a DIMM-based system since all the devices in a channel are connected to the same data and C/A buses~\cite{abcDIMM}.
As Figure~\ref{fig:broadcast} shows, we extend the {\rev{GNN}ear}-ISA and add a {B-Type} instruction. Unlike the other three instructions, a {B-Type} instruction is received and executed simultaneously by all the DIMMs in a channel.  After receiving a B-Type instruction, NMEs' controllers know that the following memory write commands are broadcast-write commands and ignore the \texttt{DIMM} fields in the address. As shown in the timing diagram, a single WR command following a B-type instruction will  write the same data to multiple DIMMs. To achieve this, we also place the duplicated data in the same area in each DIMM. Moreover, since we only duplicate high-degree vertices within each  channel, \emph{no inter-channel broadcast is needed.} The overhead of extra memory access incurred by HGP is estimated in Section~\ref{sec:performance_analysis}.
\input{figtex/fig_broadcast.tex}

\subsection{Load-balanced Interval Scheduling}
\noindent According to \texttt{\rev{GNN}ear}'s base workflow and the Narrow-Shard strategy, \texttt{\rev{GNN}ear} computes intervals sequentially ('Interval' denotes a column of shards computing the same destination vertices). 
The CAE will not send instructions for  interval $i+1$ before finishing interval $i$. 
However, the irregularity of graphs can make the assigned shards within an interval vary among DIMMs, causing load-imbalance problems. 
The main idea to tackle such a problem is to start the processing of the next intervals on those idle DIMMs. To make such an idea practical, we apply the following two techniques:

\input{figtex/fig_interval_scheduling}

\noindent\textbf{Window-based Scheduling:} To efficiently manage the concurrent intervals, we  propose a Window-based Scheduling strategy. As Figure~\ref{fig:sliding} shows, we set several result FIFOs and a window buffer in CAE. Each FIFO receives partial reduction results from a single DIMM, which will be merged with the partial results stored in the window buffer.  
We allow the CAE-side controller to issue instructions for interval $i+1$ immediately after a DIMM finishes interval $i$, if interval $i+1$ is within the \emph{Processing} window. The partial results in FIFOs will be merged with that stored in the window buffer. Once every DIMM's results of interval $i$ are merged, CAE commits interval $i$ and right-shifts the \emph{Processing} window. 
By this means, we can schedule multiple intervals concurrently and mitigate  the  load-imbalance problems caused by graphs' irregularity.

\noindent\textbf{Intervals Interleaving:}
Our HGP strategy (Figure~\ref{fig:partition}) divides an interval into two parts: the low-degree part and the high-degree part. According to the original interval index,  the high-degree parts in the first four intervals  (we assume a shard size of 1 for simplicity) will both be processed by DIMM-0, making it overloaded if we execute these intervals  concurrently.
We mitigate this problem by interleaving intervals among DIMMs. We reorganize the interval indexes and  ensure that the adjacent intervals rely on distinct DIMMs to process the high-degree parts.

\label{sec:shard_scheduling}

\subsection{Other Optimizations}

\noindent\textbf{Inter-Shard Overlapping:}
In Figure~\ref{fig:cache}, each shard needs a load (L-Type) and multiple edge calculation (C-Type) operations. 
If they are executed sequentially,
either the memory devices or the execution unit will be idle at a certain time. To improve resource utilization, we consider overlapping L-Type and C-Type instructions of the adjacent shards.
We use Step-5 and Step-6 in Figure~\ref{fig:cache} as an example. As shown in  Figure~\ref{fig:timing}, supposing the vector size is 128B, after two burst reads, $v_2$'s vector has been loaded to NME's data buffer. Then the calculation of $e_{2-3}$, $e_{2-4}$ can be launched immediately. In the meantime, the CAE-side controller issues the load instruction for Step-6, which is executed by NME concurrently.  Step-6's data loading  overlaps with Step-5's  computing.  This strategy increases the utilization of both the execution unit and DRAM devices and is easy to implement since all the operations have determined timing (Section~\ref{sec:ISA}), and we only schedule the two adjacent shards.
 
\input{figtex/fig_inter_shard_scheduling}

\input{figtex/fig_perforamance_comparsion}

\noindent\textbf{Interchange the Execution Order:}
Finally, for some tasks the input feature  can be much longer than the hidden feature. For example, Reddit's input feature length is 602, while the hidden size is usually 128 or 256.
Actually, if aggregators are linear, the execution order of aggregation and combination can be exchanged to reduce DRAM access and save NME's data buffer.
For instance, we can calculate the combination  first: $a^1_u = h_u^1\cdot W^1$, then aggregation: $h_v^2 = \sigma(\sum_{u\in \widetilde{\mathcal{N}}(v)}\frac{1}{\sqrt{D_uD_v}}\cdot a_u^1)$. 
The original execution order incurs roughly  $ 2\times |E|\times d_1 + |V|\times d_2 $ DRAM access for the first layer, where $d_1$ and $d_2$ denote the input and output feature dimensions. After interchanging the execution order, the memory access becomes  $2\times |E| \times d_2 + |V|\times d_1$. Since  $|V|<<|E|$,  the data access reduces about $\frac{d_1}{d_2}\times $. With hidden size = 256, the theoretical DRAM access reduces roughly $2.3\times$ for Reddit.

%% file: figtex/fig_degree.tex
\begin{figure} [t]
    \centering
    \includegraphics[width=1.0\linewidth]{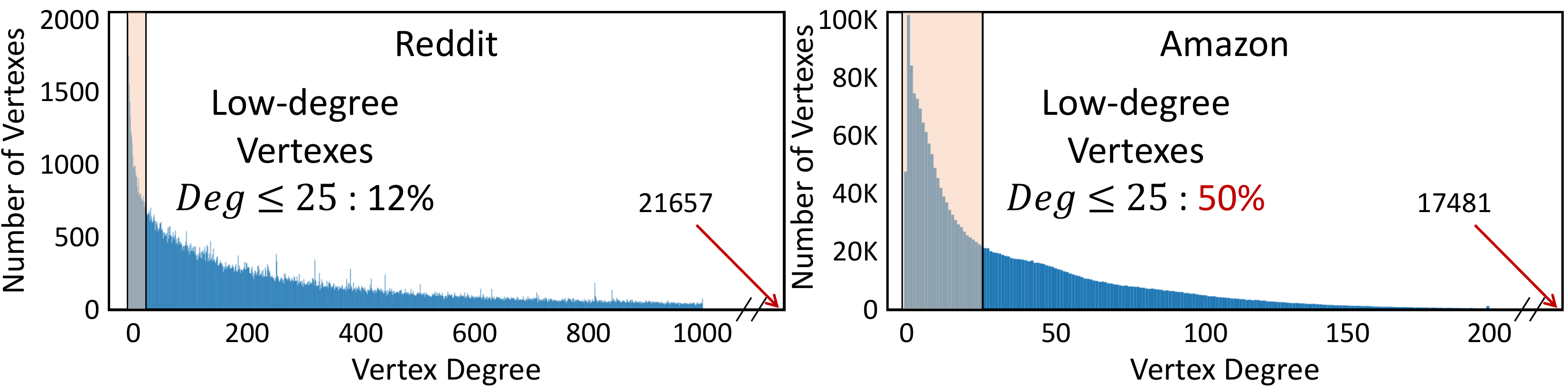} 
      \caption{The degree distribution of Reddit and Amazon.}
            \label{fig:degree}
\end{figure}

%% file: figtex/fig_partition.tex
\begin{figure} [t]
    \centering
    \includegraphics[width=1.0\linewidth]{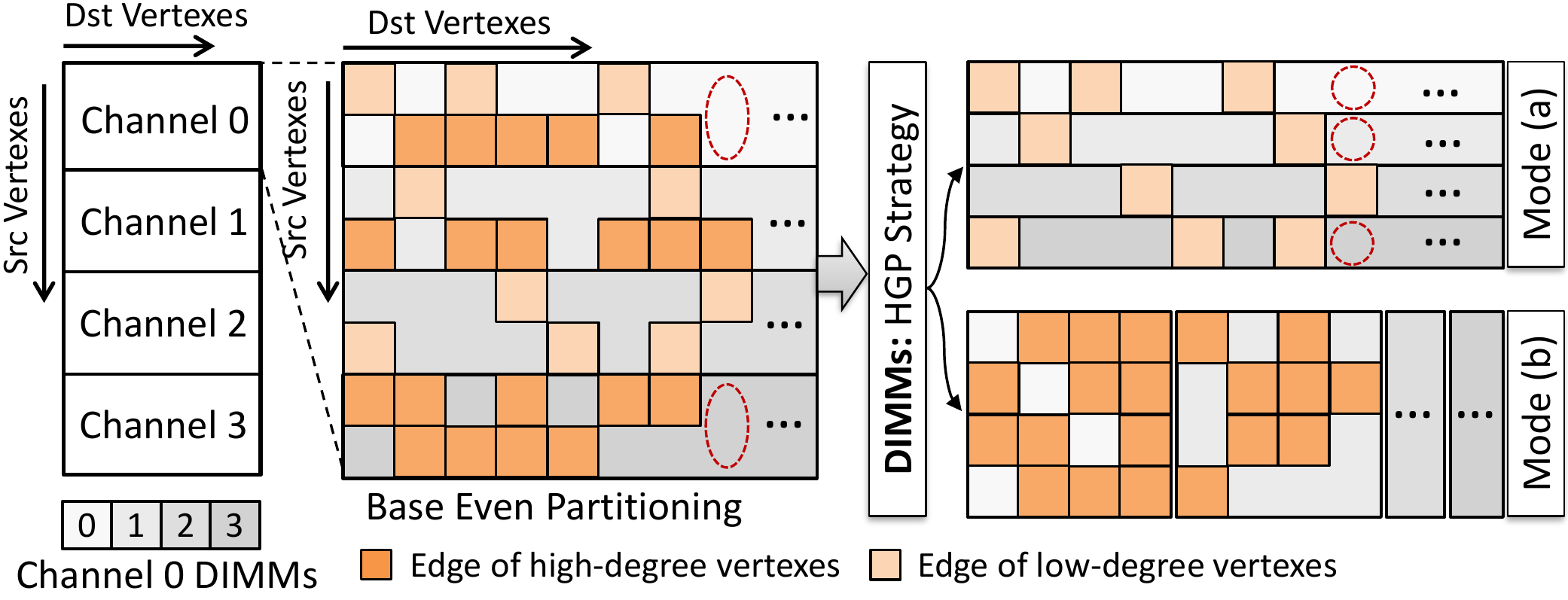}
    \caption{The HGP strategy. 
    }
         \label{fig:partition}
\end{figure}

%% file: figtex/fig_broadcast.tex
\begin{figure} [t]
    \centering
    \includegraphics[width=1.0\linewidth]{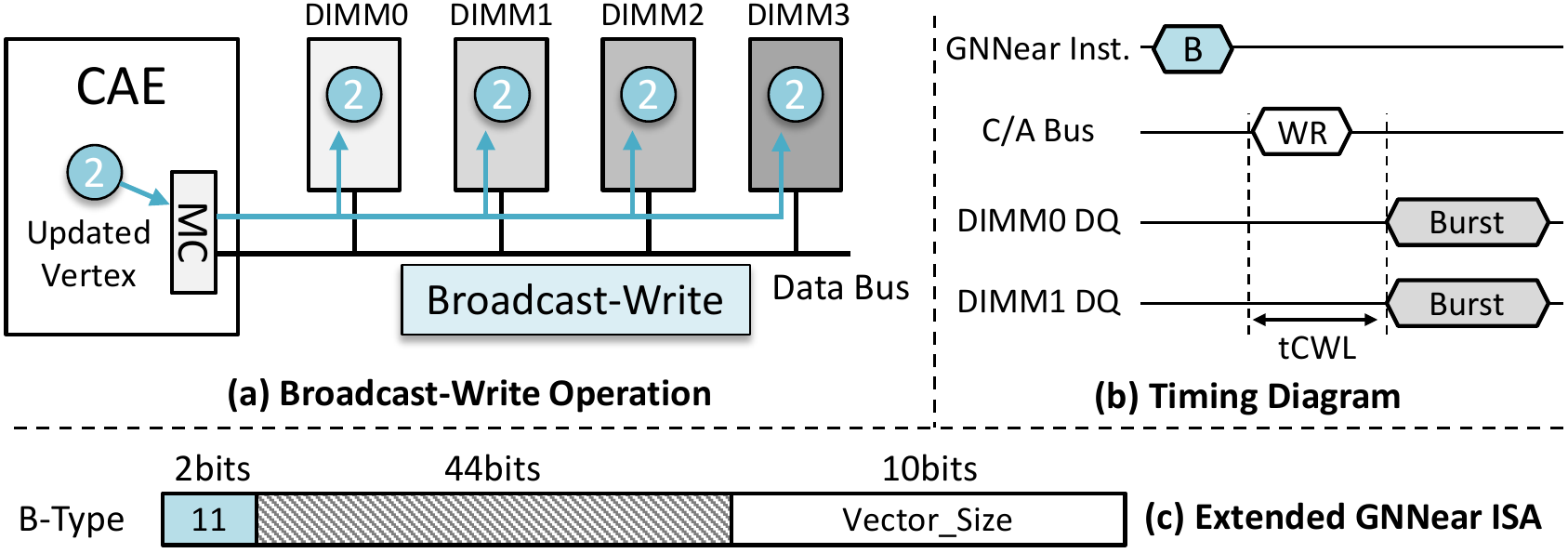} 
      \caption{The support for broadcast-write operations.}
            \label{fig:broadcast}
\end{figure}

%% file: figtex/fig_interval_scheduling.tex
\begin{figure} [t]
    \centering
    \includegraphics[width=0.98\linewidth]{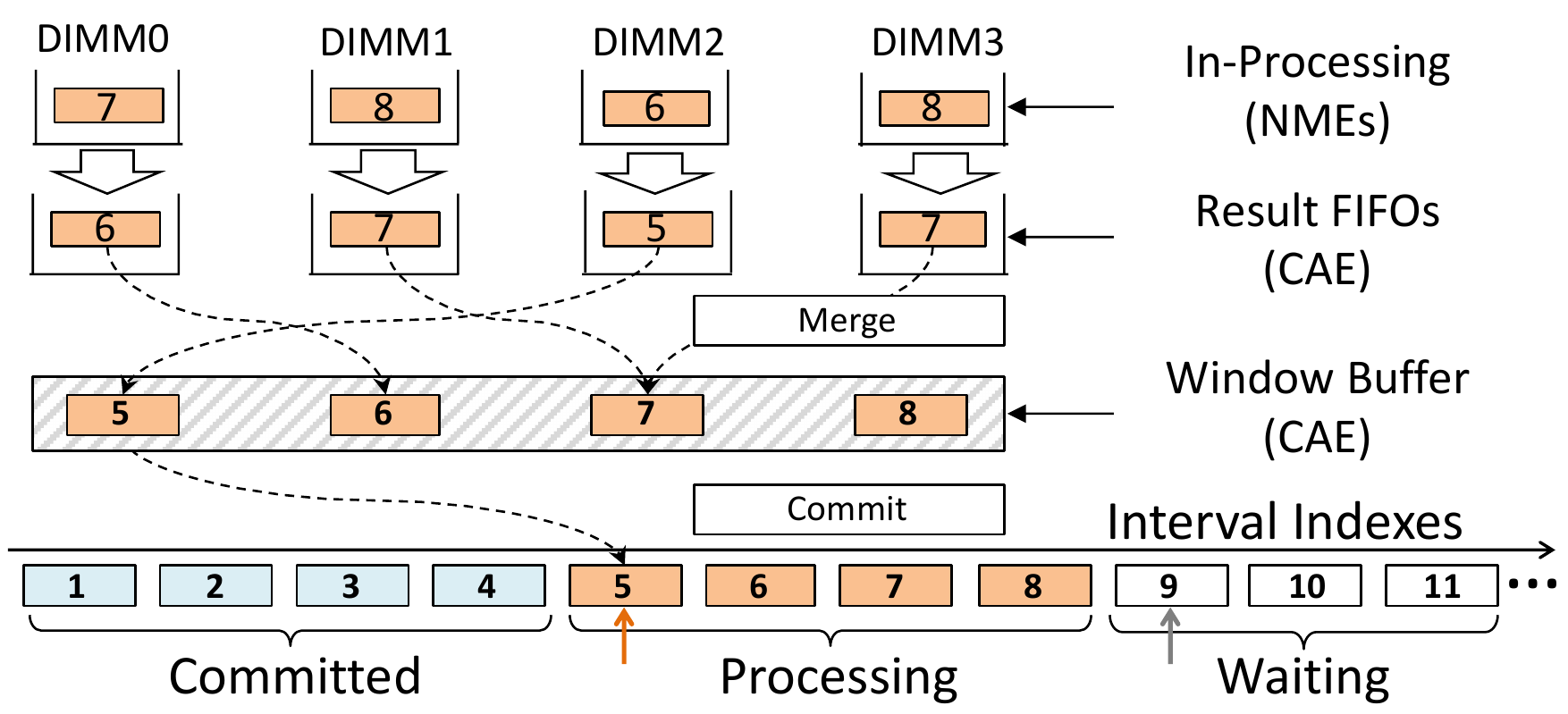} 
    \caption{Window-based scheduling.}
         \label{fig:sliding}
\end{figure}

%% file: figtex/fig_inter_shard_scheduling.tex
\begin{figure} [t]
    \centering
    \includegraphics[width=1.0\linewidth]{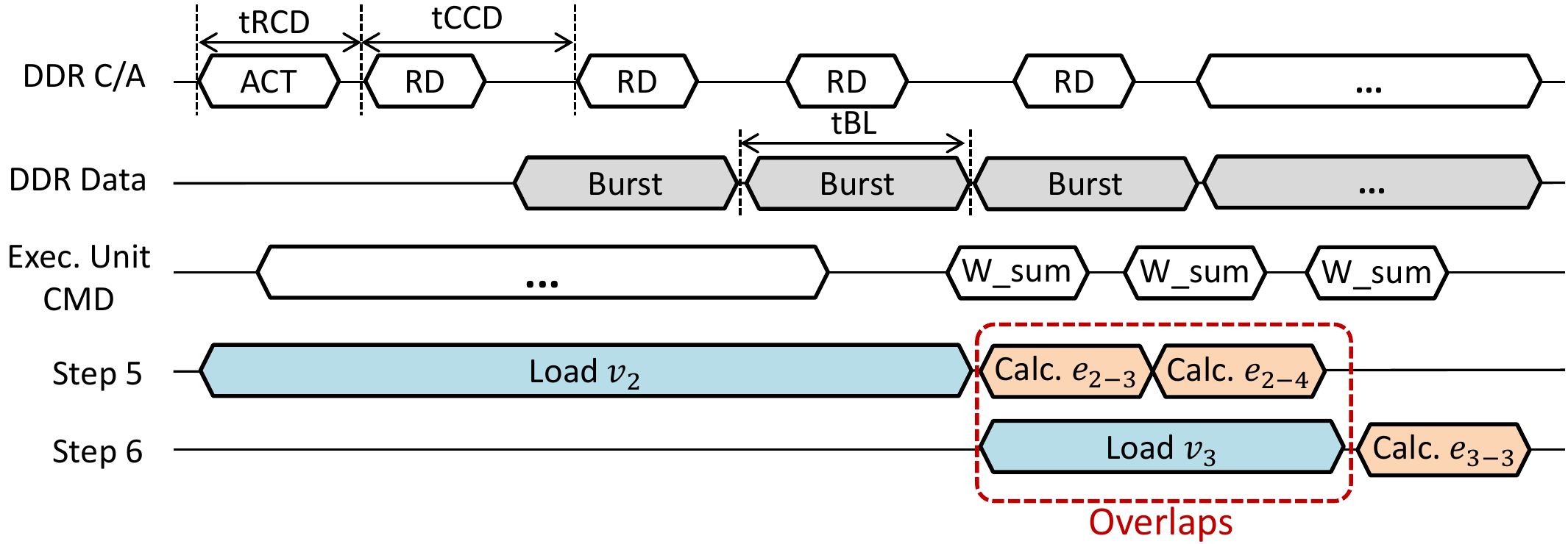} 
    \caption{Timing diagram of Inter-Shard Overlapping. }
         \label{fig:timing}
\end{figure}

%% file: figtex/fig_perforamance_comparsion.tex
\begin{figure*} [h]
\centering
    \includegraphics[width=1.0\linewidth]{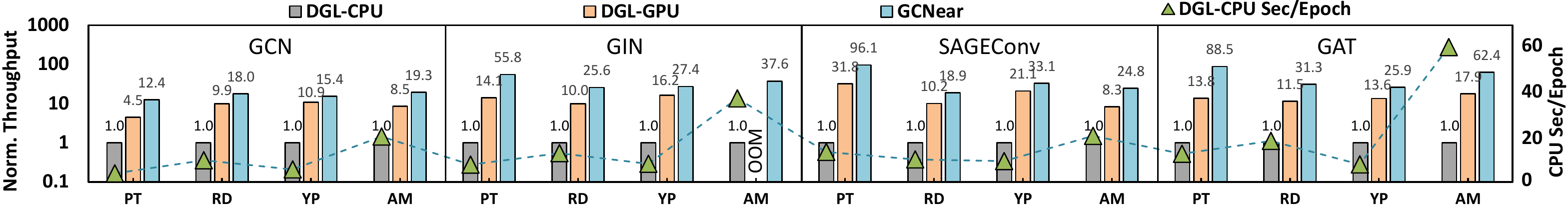} 
      \caption{Training throughput comparisons. OOM is short for ``Out of Memory''.}
            \label{performance}

\end{figure*}

%% file: tex/evaluation.tex
\section{Evaluation}
\label{sec:evaluation}
\subsection{ Methodology}
\input{figtex/fig_energy_efficiency}

\noindent\textbf{{System Configuration:}}
\input{tables/system_config}
Table~\ref{tab:config} summarizes the system parameters of the \texttt{\rev{GNN}ear} prototype. For the memory system,  each channel equips four DDR4-2400 LRDIMMs. According to Micron's LRDIMM  datasheet~\cite{LRDIMM}, each DIMM has a 32GB capacity. In total, \texttt{\rev{GNN}ear} equips 512GB of memory, which is much larger than that of V100 GPU.  
The timing setup is also based on this datasheet. 
For NME, we set 16 PEs. Each PE contains eight MACs. Running at 500MHz, an NME provides a peak performance of 128GFLOPS.
For CAE, we adopt a $128\times 128$ systolic array as the GEMM engine, providing about 22TFLOPS computation capacity running at 0.7GHz. The VPU is composed of 32 SIMD-16 cores and has 700GFLOPS peak performance. The scratchpad memory is 16MB, which can be flexibly divided into weight buffer, input/output buffer, edge buffer, window buffer, etc. We adopt  BF16 as the data format. BF16 has the same accuracy as FP32 for NN training~\cite{bf16} but is more cost-efficient. We estimate CAE and NME's area and power using 16nm and 28nm technologies, respectively. The GEMM and VPU's area and power are measured according to TPU-v2~\cite{jouppi2020domain,ten_lessons}, and HyGCN~\cite{hygcn}.  To estimate the overhead of NME's logic parts, we write EU with Chisel and synthesize the generated RTL using Synopsys Design Compiler 2016 under TSMC 28nm. We get the area and power of all SRAM buffers with CACTI~\cite{CACTI}. The main components of NME incur moderate area ($0.86$ mm$^2$) and power ($258.1$mW) overhead, given that a single DIMM usually consumes several watts and the buffer chip takes up about $100$ mm$^2$~\cite{buffer_chip}. 

\input{tables/benchmarks}

\noindent\textbf{Simulation Methodology:} To evaluate the performance of \texttt{\rev{GNN}ear}, we implement a customized \rev{GNN} training framework to support our narrow-shard-based training dataflow given a graph's adjacent matrix and the model's configuration. It generates \texttt{Reduce}, \noindent\texttt{Update}, and other operations and sends them to a  \texttt{\rev{GNN}ear} simulator. Specifically, we extend  DRAMsim3 simulator~\cite{dramsim} to support cycle-accurate near-memory reduction simulation. It is customized to support \rev{GNN}ear-ISA and takes NME's timing constraints and power consumption into consideration. Note that the DRAMSim3 simulator is too slow to evaluate large graphs which even have billions of edges. Therefore we also develop a coarse-grained simulator for quick evaluation, whose key parameters (i.e., the per-operation latency, bandwidth utilization, etc.) are derived from the fine-grained simulator.  The CAE and VPU cycles are calculated according to their parallelism and the size of features. We model on-chip buffers' access latency using CACTI~\cite{CACTI} and inject these parameters into our simulator.

\noindent\textbf{Benchmarks:}~Table~\ref{tab:graph} lists four benchmarking graphs, namely Ogbn-Proteins (PT)~\cite{proteins}, Reddit (RD)~\cite{graphSAGE}, Yelp (YP)~\cite{yelp}, and Amazon (AM)~\cite{clustergcn}.  PT and RD are dense graphs, while YP and AM are relatively sparse. We adopt four  aforementioned GCN models as benchmarking models. 
In the default configuration, each model has two layers with a hidden size of 256.

\noindent\textbf{Baselines:} Since there is no other full-batch training accelerator at present, we mainly compare \texttt{\rev{GNN}ear} with CPU/GPU platforms using the popular DGL framework~\cite{DGL}. We adopt a DGX-1 workstation equipping Xeon E5-2698-v4 CPU (4 Channels, 256GB DDR4-2400) and V100 32GB GPUs to evaluate the performance of DGL-CPU (MKL-2020.2) and DGL-GPU (CUDA-10.2/cudnn 7.6).
Considering that \texttt{\rev{GNN}ear} adopts the BF16 data format (two Bytes per element) while the CPU/GPU baselines all use FP32 (four Bytes per element),  we {double the vector size} when simulating \texttt{\rev{GNN}ear}'s data transmission for  fair  comparisons. 

\subsection{Main Results}

\noindent\textbf{Training Throughput:} We first estimate the training throughput of \texttt{\rev{GNN}ear} and the CPU/GPU baselines.  
The performance is measured with end-to-end training seconds per epoch and then normalized to the CPU baseline. As depicted in Figure~\ref{performance},  DGL-GPU suffers \rev{from} the OOM problem on the AM-GIN task due to GPU's limited memory capacity.  Though DGL-CPU successfully tackles all tasks with 256GB memory, it shows low training speed due to its limited memory bandwidth and computation capacity. \texttt{\rev{GNN}ear} demonstrates superior training speed compared to these two baselines. Specifically, \texttt{\rev{GNN}ear} achieves  \geomeanSpeedupOverCPU~and \geomeanSpeedupOverGPU~geomean speedup compared to DGL-CPU and DGL-GPU, respectively.

\noindent\textbf{Energy-efficiency:}
We measure \texttt{\rev{GNN}ear}'s  per-epoch energy consumption and compare it with CPU/GPU baselines. The CPU system's energy is directly estimated with PyRAPL~\cite{pyRAPL}.  We test the GPU's running power using PyNVML~\cite{PyNVML} and then calculate the energy with the product of average power and per-epoch time. All values are also normalized to the CPU baseline.  As shown in Figure~\ref{fig:energy-eff}, \texttt{\rev{GNN}ear} achieves \geomeanEnergySavingOverCPU~and \geomeanEnergySavingOverGPU~higher energy efficiency (geomean) compared to CPU/GPU platforms. High training throughput and low power consumption brought by the ASIC engine and saved data transmission with near-memory reduction jointly lead to \texttt{\rev{GNN}ear}'s extraordinary energy efficiency.

\subsection{Performance Analysis}
\label{sec:performance_analysis}

\noindent\textbf{Speedup Breakdown:} To better understand the effect of different design points, we demonstrate the speedup breakdown in  Figure~\ref{fig:breakdown}-(a). Evaluations are performed on AM-GCN task  \rev{and the performance is normalized to the CPU baseline}. As we can see, directly adopting CAE for \rev{GNN} training gets $3.6\times$ speedup, thanks to CAE's  higher performance over CPU. Performing near-memory reduction  brings $3.8\times$ speedup via utilizing the high aggregated in-DIMM bandwidth. Adopting the Narrow-Shard strategy to explore data reuse brings $1.1\times$ speedup \rev{thanks to the reduced local DRAM access}. The HGP strategy and  load-balanced interval scheduling further contribute to $1.2\times$ speedup.  At last, inter-shard overlapping contributes to $1.1\times$ speedup further by mitigating the idle time of NMEs. Note that for different tasks, these proposed optimizations show diverse impacts on \texttt{\rev{GNN}ear}'s performance (for instance, RD graph gets much higher speedup from the Narrow-Shard strategy). Since AM does not need to interchange the execution order to save memory access, we evaluate the Interchange Execution Order (IEO) strategy on RD. As shown in Figure~\ref{fig:breakdown}-(b), IEO successfully brings $1.5\times$ speedup on the RD-GCN task thanks to the reduced memory access in the first layer.  Generally, our designs and optimizations demonstrate significant effectiveness in accelerating \rev{GNN} training. 

\input{figtex/fig_breakdown}

\input{figtex/fig_roofline}

\noindent\textbf{Roofline Analysis: }
We adopt a roofline model to analyze the performance of \texttt{\rev{GNN}ear} and DGL-CPU baseline on the GIN model. 
As shown in Figure~\ref{fig:roofline}-(a), \rev{the X-axis represents the operational intensity, while the Y-axis is the performance (both in log scale).} The Xeon E5-2698-v4 CPU and \texttt{\rev{GNN}ear} both have four memory channels, providing 76.8GB/s bandwidth. \rev{We plot four operations using this model: CPU-Update, CPU-Reduce, GNNear-Update and GNNear-Reduce. As we can see,} CPU-Reduce suffers from the low arithmetic intensity and is bounded by the limited memory bandwidth (\rev{the left-most triangle}). In comparison, \texttt{\rev{GNN}ear}'s near-memory reduction mechanism provides up to $8\times$ higher aggregated local bandwidth. Moreover,  the Narrow-Shard strategy increases the arithmetic intensity of \texttt{Reduce} operations. Therefore, \rev{GNN}ear-Reduce achieves  $14.5 \times$ higher performance compared to CPU-Reduce. Besides, \rev{GNN}ear-Update also shows $25.4 \times$ speedup against CPU-Update due to the more powerful CAE and higher arithmetic intensity by buffering all the weights on-chip.

\noindent\textbf{DRAM Access Saving:} By performing near-memory reduction, the data read via memory channels is substantially reduced. \rev{According to our profiling,} the evaluated graphs \rev{can} benefit from  $71.3\%$ to $97.2\%$ off-chip memory-read saving \rev{by adopting near-memory processing}, which dramatically reduces the system's energy consumption. Assuming the off-chip IO cost is 22pJ/b, and the on-chip DRAM read cost is 14pJ/b~\cite{recnmp}, about $43.6\%$$\sim$$59.4\%$ of total data read energy is saved. \rev{We illustrate the number of read/write instructions of AM-GCN task under different settings (without NMP, NMP without broadcast write, NMP with broadcast write) in Figure~\ref{fig:roofline}-(b). As we can see},   on the representative AM-GCN task, the HGP strategy incurs $1.05\times$ extra off-chip DRAM write with a $\lambda$ of $0.35$ (\rev{The orange bar in the middle of} Figure~\ref{fig:roofline}-(b)). Fortunately, as shown by \rev{the NMP+BW column in Figure~\ref{fig:roofline}-(b)}, the proposed broadcast write mechanism eliminates the extra off-chip data write and ensures that the write-back operations will not be the bottleneck.  Therefore, the  energy cost incurred by broadcast-write operations is merely about $8.4\%$ of memory access energy.

\subsection{Design Space Exploration}
\label{sec:dse}

\noindent\textbf{Shard Size \& Window Size: }  
\input{figtex/fig_dse}
To understand the impact of shard size and window size on \noindent\texttt{\rev{GNN}ear}'s performance,  we first keep $window$ = 4 and set $shard$ from 1 to 256, and evaluate the training speedup on GCN tasks. As Figure~\ref{fig:DSE}-(a) shows, RD and PT are more sensitive to shard size, while the speedup on AM and YP is quickly saturated as the shard size increases. Considering that a large shard demands much more NMEs' data buffer, we set $shard$ = $128$ in our prototype. We then explore $window$  from $1$ to $32$.  In Figure~\ref{fig:DSE}-(b), RD benefits a lot from a large window. The speedup on all the tasks increases first and then reaches a plateau.  Since the space complexity of window buffer is $shard \times  window \times d$, it is better to set the window size to a relatively small value (e.g., $window$ = $4$ in our prototype) to save CAE's area and energy.

\noindent\textbf{Duplication Ratio:} We explore HGP's duplication ratio $\lambda$ from $0$ to $0.5$. As depicted in Figure~\ref{fig:DSE}-(c), YP and AM obtain considerable speedup from duplicating high-degree vertices. However, RD is not sensitive to $\lambda$, and PT even gets a lower speed. We infer that this is because PT and RD are dense graphs and do not have enough low-degree vertices (as Figure~\ref{fig:degree} shows, in RD, only 12\% of vertices have degrees $\leq 25$). The baseline even-partitioning strategy already works well. Duplicating vertices only increases their load imbalance.  Therefore, we choose not to duplicate vertices on PT and RD and tentatively set $\lambda = 0.35$ on AM and YP.

\noindent\rev{\textbf{Ranks Per DIMM:} An NME can access ranks in parallel to achieve \emph{\#Rank}$\times$ higher local bandwidth. To study the benefits of rank-level parallelism, in Figure~\ref{fig:DSE}-(d), we explore the number of  ranks  per DIMM from 1 to 8. The speedup will also  be saturated when near-memory reduction is bounded by the channel bandwidth or NME's computation capacity. Adding multiple ranks in a DIMM and driving them in parallel will  increase the complexity of NME's interface and control logic. Therefore we consider two ranks per DIMM in our prototype.}

%% file: figtex/fig_energy_efficiency.tex
\begin{figure*} [t]
\centering
    \includegraphics[width=1.0\linewidth]{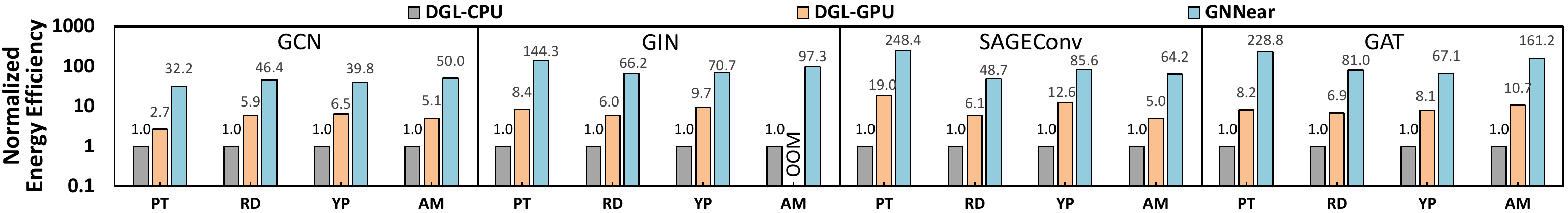} 
      \caption{Energy-efficiency comparisons. OOM is short for ``Out of Memory''.}
            \label{fig:energy-eff}

\end{figure*}

%% file: tables/system_config.tex
\begin{table}[t]
\caption{System Parameters and Configurations}
\label{tab:config}
\centering
\setlength{\tabcolsep}{2.0mm}{
\resizebox{0.48\textwidth}{!}{
\renewcommand{\arraystretch}{0.75}{
\begin{tabular}{ccl}
\toprule
\multicolumn{3}{c}{\textbf{Memory System Configuration}}                                                                                                                                                                                                                \\ \midrule
\multicolumn{3}{c}{\begin{tabular}[c]{@{}c@{}}DDR4-2400, 32GB LRDIMM, 4 channels $\times$ 4 DIMMs $\times$ 2 ranks \\ FR-FCFS, 32 entry RD/WR queue, Open policy\end{tabular}}                                                                                  \\ \midrule
\multicolumn{3}{c}{\textbf{DRAM Timing Parameters}}                                                                                                                                                                                                                     \\ \midrule
\multicolumn{3}{c}{\begin{tabular}[c]{@{}c@{}}tRC=56,  tRCD=17, tCL=17, tRP=17, tBL=4\\ tCCD\_S=4, tCCD\_L=6,tRRD\_S=4, tRRD\_L=6, tFAW=26\end{tabular}}                                                                                                               \\ \midrule
\multicolumn{3}{c}{\textbf{NMP Parameters}}                                                                                                                                                                                                                             \\ \midrule
\multicolumn{1}{c|}{Data  Buffer}                                   & \multicolumn{2}{c}{\begin{tabular}[c]{@{}c@{}}256KB,  Dual Ports, Word size = 16B \\ Area = $0.44$ mm$^2$, Power = $80.0$ mW\end{tabular}}                 \\ \midrule
\multicolumn{1}{c|}{Execution  Unit}                                & \multicolumn{2}{c}{\begin{tabular}[c]{@{}c@{}}16 PEs, 8 MACs per PE @500MHz\\ Area = $0.42$ mm$^2$, Power = $178.1$ mW\end{tabular}} \\ \midrule 
\multicolumn{3}{c}{\textbf{CAE Parameters}}                                                                                                                                                                                                                             \\ \midrule
\multicolumn{1}{c|}{GEMM  Engine}                                   & \multicolumn{2}{c}{\begin{tabular}[c]{@{}c@{}} 128$\times$128 Systolic Array @700MHz\\  Area = 27.3 mm$^2$,  Power = $ 6291.4 $ mW\end{tabular}}             \\ \midrule
\multicolumn{1}{c|}{\begin{tabular}[c]{@{}c@{}}VPU\end{tabular}}                                   & \multicolumn{2}{c}{\begin{tabular}[c]{@{}c@{}} SIMD-16 Cores $\times 32$ @700MHz\\  Area = 0.82 mm$^2$,  Power = $296.6$ mW\end{tabular}}             \\ \midrule
\multicolumn{1}{c|}{Scratchpad Memory}                              & \multicolumn{2}{c}{\begin{tabular}[c]{@{}c@{}}16MB, 8 Banks, Dual Ports, Word size = 64B \\ Area= 33.7 mm$^2$, Power = $5519.2$ mW\end{tabular} }                                                                           \\ \bottomrule 
\end{tabular}}}}
\end{table}

%% file: tables/benchmarks.tex
\begin{table}[t]
\centering
\caption{Graph Datasets}
\label{tab:graph}
\footnotesize
\setlength{\tabcolsep}{1.2mm}{
\renewcommand{\arraystretch}{0.8}{
\begin{center}
\resizebox{0.48\textwidth}{!}{
\begin{tabular}{c|c|c|c|c}
\toprule
Graph       & \#Vertexes & \#Edges &\#Features   & Avg. Degree   \\ 
\midrule
Ogbn-Proteins (PT) & 132,534 & 39,561,252 & 128 &   597.0  \\
Reddit (RD) & 232,965 & 114,615,892 & 602 &     492.9       \\ 
Yelp (YP)   & 716,847    &  6,977,410     & 300 &   19.5       \\
Amazon (AM) &  2,449,029  &     123,718,280  &  100    &    101.0   \\
\bottomrule
\end{tabular}}
\end{center}
}
}
\end{table}

%% file: figtex/fig_breakdown.tex
\begin{figure} [t]
    \centering
    \includegraphics[width=1.0\linewidth]{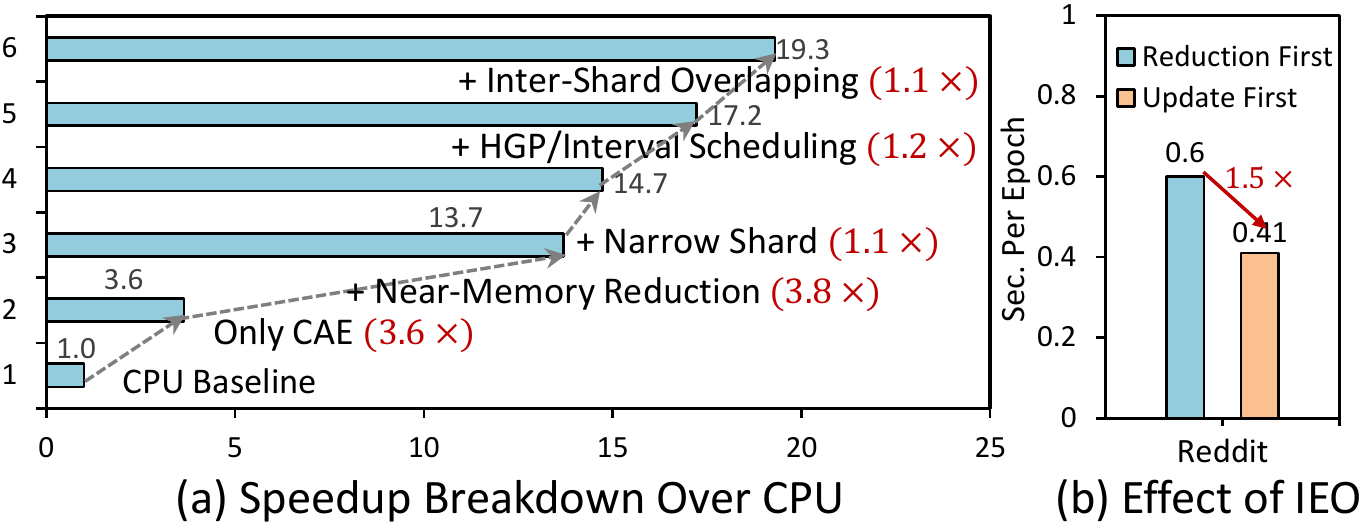} 
      \caption{Performance analysis.}
            \label{fig:breakdown}
\end{figure}

%% file: figtex/fig_roofline.tex
\begin{figure} [t]
    \centering
    \includegraphics[width=1.0\linewidth]{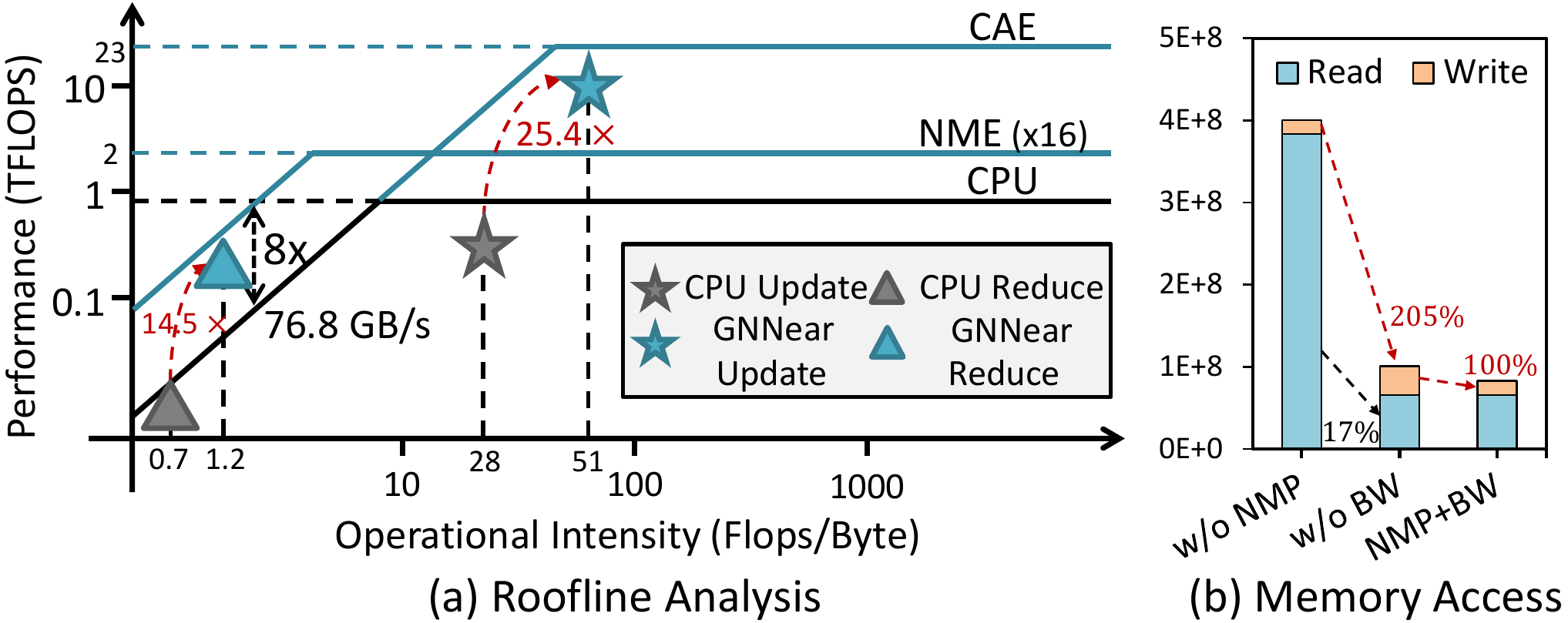} 
      \caption{Roofline analysis and memory access breakdown. 
      BW is short for the Broadcast-Write mechanism.
      }
            \label{fig:roofline}
\end{figure}

%% file: figtex/fig_dse.tex
\begin{figure} [t]
    \centering
    \includegraphics[width=1.0\linewidth]{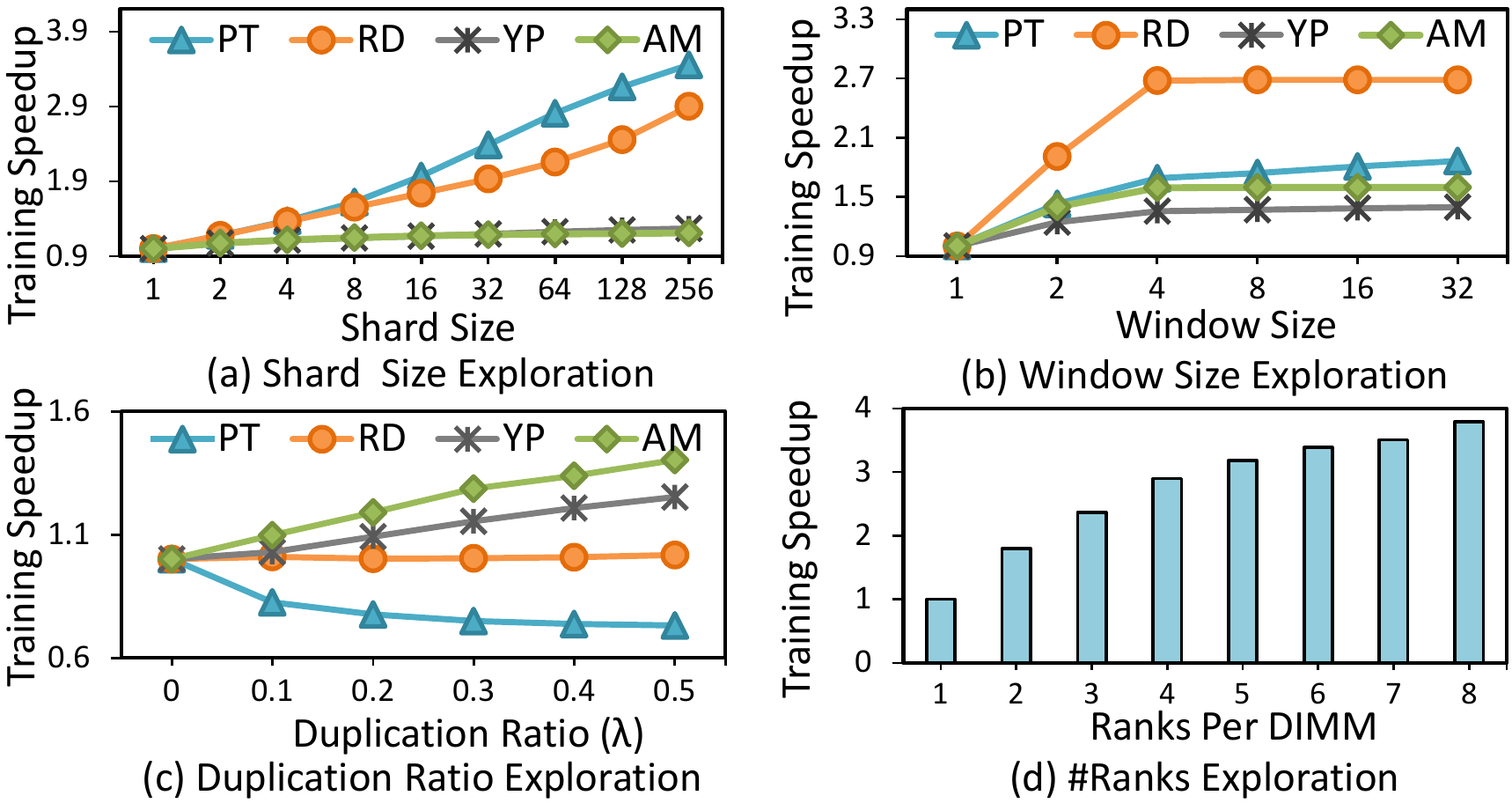} 
      \caption{Design space exploration with the GCN model. Experiment (d) is based on the AM dataset.}
            \label{fig:DSE}
\end{figure}

%% file: tex/discussion.tex

\subsection{Comparisons with Roc and DistGNN}
\label{sec:multi}

\noindent We notice that there are some distributed GPU/CPU-based  full-batch \rev{GNN} training systems, such as Roc~\cite{roc} and DistGNN~\cite{distgnn}, which  support deep \rev{GNN} models (more than two layers) and even super large graphs like Ogbn-Papers~\cite{hu2020ogb}. To compare with them,  we add routers to CAE and connect multiple $\texttt{\rev{GNN}ear}$ accelerators with a switch and also build a Multi-\rev{GNN}ear system. As shown in Figure~\ref{fig:multi}, we partition the graph evenly and assign the sub-graphs to different  accelerators (four \texttt{\rev{GNN}ear} accelerators are named  G-0 to G-3). Each accelerator further partitions the sub-graph to its connected DIMMs.  We assign \texttt{Reduce} and \texttt{Update} operations concerning different destinations to each accelerator. Moreover, accelerators transmit locally-merged partial results to each other to save inter-chip communication.  
\input{figtex/fig_multi_gcnear}

\input{tables/distributed_system_configuration}

 We evaluate the performance of Multi-\rev{GNN}ear by extending our simulator.  Table~\ref{tab:dist_param} lists the parameters of the three systems. We assume the switch of Multi-\rev{GNN}ear has the same bandwidth as Roc's NVLink. We adopt the same model and graph settings from their papers  and  use their reported performance numbers  for comparison.  In general, Multi-\rev{GNN}ear achieves about  $3.1\times$ speedup on  Ogbn-Papers (OP) dataset (3-layer GCN, hidden size = 256) and is also $1.6\times$ faster on AM than DistGNN with 32 CPU sockets.   On deep model tasks (four-layer GCN), Multi-\rev{GNN}ear achieves about $2.1\times$ speedup on AM dataset and $1.08\times$ speedup on the RD dataset, compared to the Roc system built with eight V100 GPUs. 

\subsection{Comparisons with Rubik and GraphACT}

\noindent\textbf{Comparison with Rubik:} Rubik~\cite{rubik} \rev{can also be used for} GCN training. The main idea of Rubik is  using LSH hashing to reorder the input graphs for better data locality. 
However, the effect of graph reordering purely  depends on the pattern of graphs. The heavy pre-processing overhead (more about ten seconds on Reddit) also restricts its adoption to offline applications~\cite{geng2021gcn}. \texttt{\rev{GNN}ear} reduces DRAM access through near-memory processing which is a more generic solution. The pre-processing step required by HGP is more than $5\times$ faster than Rubik, according to our evaluation. Such a light-weighted  pre-processing  can even be omitted  when training on dense graphs (see Section~\ref{sec:dse}).

\noindent\textbf{Comparison with GraphACT:} GraphACT is the state-of-the-art mini-batch GCN training accelerator, which is implemented on FPGAs. With the proposed GraphSAINT~\cite{graphsaint} mini-batch training algorithm, it reports $95.2\%$ accuracy on the Reddit dataset. However, for full-batch training, we can achieve $96.9\%$ accuracy~\cite{roc}, about 1.7 points higher than mini-batch training.  GraphACT cannot support full-batch training on large graphs due to the limited memory capacity of FPGAs. Therefore, \texttt{\rev{GNN}ear} can easily outperform GraphACT in accuracy. Moreover, unlike \texttt{\rev{GNN}ear} adopting HGP strategy, GraphACT can hardly handle low-degree graphs~\cite{rubik}.

%% file: figtex/fig_multi_gcnear.tex
\begin{figure} [t]
    \centering
    \includegraphics[width=0.98\linewidth]{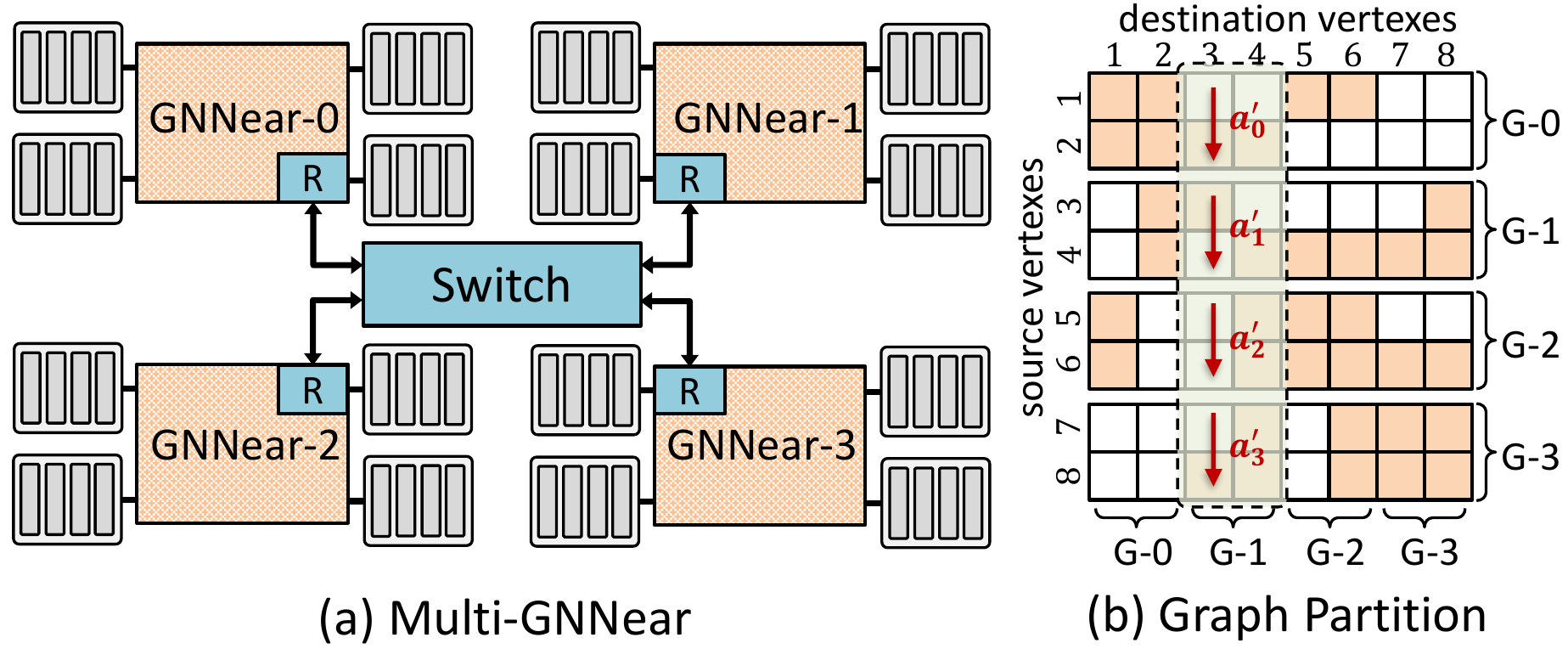} 
    \caption{Multi-{GNN}ear system and the graph partition. }
         \label{fig:multi}
\end{figure}

%% file: tables/distributed_system_configuration.tex
\begin{table}[t]
\caption{Parameters of Distributed Systems}
\label{tab:dist_param}
\setlength{\tabcolsep}{1.1mm}{
\resizebox{0.48\textwidth}{!}{
\renewcommand{\arraystretch}{1.2}{
\centering
\begin{tabular}{c|c|c|c}
\toprule
             & Configuration                        & Total Memory &  Computation Capacity \\ \midrule
Roc~\cite{roc}         & 8 NVIDIA-V100, 2 nodes                & 256GB        & 126 TFLOPS                 \\ \midrule
DistGNN~\cite{distgnn}      & 32 Intel Xeon 9242 CPU & \textgreater~2TB        & 226 TFLOPS                 \\\midrule
Multi-\rev{GNN}ear & 4 \rev{GNN}ear accelerator                 & 2TB          & 100 TFLOPS                 \\ \bottomrule
\end{tabular}}}}
\end{table}

%% file: tex/related_work.tex
\section{Related Work}
\noindent\textbf{\rev{GNN} Acceleration:}
Recently, plenty of GNN  accelerators have been presented \cite{hygcn, geng2019awb, blockgnn, liang2020engn, rubik, legognn,gcnx,song2021cambricon,zhang2020hardware,wang2020gnn,nie2021gcn} for efficient \rev{GNN} inference. As far as we know,
GraphACT~\cite{graphact} is the only \rev{GNN} training accelerator but it just supports mini-batch training on middle-scale graphs like Reddit and Yelp. 
Our \texttt{\rev{GNNear}} accelerator targets the more accurate but challenging large-scale full-batch training tasks. Besides, there are also several  mini-batch/full-batch training frameworks~\cite{dgcl, roc, distgnn, neugraph, tripathy2020reducing, aligraph,fuseGNN,p3,dorylus,mohoney2021marius}. They  are based on general-purpose CPU/GPU platforms and cannot benefit from domain-specific hardware.

\noindent\textbf{DRAM-based Near-Memory Processing:}
Many near-memory processing accelerators using 3D/2.5D-stacked memory have been proposed for graph processing~\cite{zhuo2019graphq,graphp,ahn2015scalable, dai2018graphh,graph_pim}, DNN acceleration~\cite{gao2017tetris,WangCYL18,deep_train,liu2018processing,HongRK18,LeeH21,Parana,SchuikiSGB19,HBM2}, GCN inference~\cite{chen2021towards} or general-purpose applications~\cite{spaceA, gu2020ipim,top_pim,NDA,GaoAK15,hsieh2016transparent}. Due to the limited memory capacity ($\leq$128GB) and high cost, 3D/2.5D-stacked NMP is not suitable for full-batch \rev{GNN} training.  Chameleon~\cite{chameleon} proposes to adopt  LRDIMMs~\cite{LRDIMM} to break NMP's capacity limitation. Several follow-up  works adopt this  paradigm to build efficient recommendation systems~\cite{tensordimm, recnmp,park2021trim,FAFNIR,ke2021near} or accelerate extreme classification~\cite{enmc}. Recently, Samsung has also  disclosed a concept DIMM-NMP product called AXDIMM~\cite{AXDIMM}, which still adopts recommendation system as its killer application.
Our work is the first to leverage DIMM-NMP to accelerate full-batch \rev{GNN} training.

%% file: tex/conclusion.tex
\section{Conclusion}

\noindent In this paper we propose \texttt{\rev{GNNear}}, a hybrid accelerator  architecture leveraging  near-memory processing to accelerate full-batch \rev{GNN} training on large graphs. To deal with the irregularity of graphs, we also propose several optimization strategies concerning data reuse, graph partitioning, and dataflow scheduling, etc.  Evaluations on 16 tasks demonstrate that \texttt{\rev{GNN}ear} achieves \geomeanSpeedupOverCPU~/~\geomeanSpeedupOverGPU~geomean speedup and \geomeanEnergySavingOverCPU~/ \geomeanEnergySavingOverGPU~(geomean) higher energy efficiency  compared to Xeon E5-2698-v4 CPU and V100 GPU. 

\section*{Acknowledgment}
\noindent We thank all the reviewers for their valuable comments.
This work is supported by NSF of China (61832020, 62032001, 92064006), Beijing Academy of Artificial Intelligence (BAAI), and 111 Project (B18001).